\documentclass[11pt]{article}

\usepackage[preprint]{acl}
\usepackage{times}
\usepackage{latexsym}
\usepackage[T1]{fontenc}
\usepackage[utf8]{inputenc}
\usepackage{microtype}
\usepackage{inconsolata}
\usepackage{booktabs}
\usepackage{amsfonts}
\usepackage{amsmath}
\usepackage{graphicx}
\usepackage{siunitx}
\usepackage{csquotes}
\usepackage{enumitem}
\usepackage{placeins}
\title{The Shared Latent Structure of LLM Psychometric Responses}

\author{
  Hubert Plisiecki\textsuperscript{1}, Sabina Siudaj\textsuperscript{2}, Kacper Dudzic\textsuperscript{1} \\
  \textbf{Anna Sterna\textsuperscript{1}, Maciej G\'{o}rski\textsuperscript{2}, Karolina Dro\.{z}d\.{z}\textsuperscript{1}} \\
  \textbf{Marcin Moskalewicz\textsuperscript{1}} \\
  \textsuperscript{1}IDEAS Research Institute \\
  \textsuperscript{2}University of Warsaw \\
  \texttt{hplisiecki@gmail.com}
}
\hypersetup{
  pdftitle={The Shared Latent Structure of LLM Psychometric Responses},
  pdfauthor={Hubert Plisiecki, Sabina Siudaj, Kacper Dudzic, Anna Sterna, Maciej G\'{o}rski, Karolina Dro\.{z}d\.{z}, Marcin Moskalewicz}
}

\begin{document}

\maketitle

\begin{abstract}
Human psychometric questionnaires are increasingly used to characterize large
language models (LLMs), but it remains unclear what exactly their responses reflect. 
To answer this question we analyze responses from 50 LLMs across 45
questionnaires spanning multiple psychological domains using psychometric methodology. 
We found a dominant principal component explaining 47.1\% of variance across standardized, response-derived
questionnaire-level primary scores, revealing substantial shared structure across instruments.
We investigate its meaning through semantic analysis, respondent framing, and
human annotations. Semantic analysis suggests that the component partly reflects
experiential self-attribution: how readily models endorse statements describing
subjective states and bodily experiences. A model score emphasizing items whose
disagreement decreases under human simulation correlates strongly with the
component ($r=.864$). Human ratings of 289 matched items show that inner-life
content is positively associated, and social desirability negatively associated,
with item--component correlations after mutual adjustment and controls for other
semantic properties.
Together, six rated properties explain 22.1\% of variance in Fisher-transformed
item--component correlations, supporting a partial interpretation involving both
experiential self-attribution and socially desirable responding. These findings
suggest that LLM psychometric profiles share response tendencies that require
interpretation beyond the human traits their questionnaires were designed to
measure.
\end{abstract}

\section{Introduction}
\label{sec:intro}

Large language models (LLMs) are increasingly assessed using human questionnaires.
Given the textual interface of these systems, applying established psychometric questionnaires to LLMs 
becomes a natural research direction.
A rapidly growing literature has used established human instruments to study LLM personality, values, morality, ideology, 
and related individual differences \citep{abdulhai_moral_2024,bodroza_personality_2024,kamal_detailed_2025,miotto_who_2022,pellert_ai_2024,sakhawat_political_2026}. 
More broadly, recent reviews suggest that 
“LLM psychometrics” is now an identifiable research area, spanning self-assessment, value measurement, personality assessment, 
and psychometric evaluation more generally \citep{brito_modeling_2025,wen_self-assessment_2024,ye_large_2026}. 

Existing work broadly treats LLMs either as simulators of human or
persona-conditioned responses
\citep{jiang_personallm_2024,serapio-garcia_psychometric_2025,wang_evaluating_2025},
or as the targets of assessment themselves
\citep{abdulhai_moral_2024,bodroza_personality_2024,li_decoding_2025,pellert_ai_2024,sakhawat_political_2026}.
It has shown that LLMs can generate stable, generalizable psychological profiles, be steered toward 
distinct psychometric personas, and sometimes be meaningfully differentiated using questionnaires or psychologically motivated benchmarks 
\citep{han_value_2025,jiang_personallm_2024,lee_llms_2025,serapio-garcia_psychometric_2025}. However, interpreting their scores as measurements of the corresponding human constructs requires validation. Recent work has shown that LLM answers do not reproduce the same latent structure we would
expect from a sample for which these tools were created \citep{suhr_challenging_2024}. Furthermore, questionnaire scores can shift 
substantially under changes in prompt wording or fine-tuning regime, raising further 
doubts about whether these instruments are measuring stable underlying constructs in LLMs at all \citep{kamal_detailed_2025}.
Taken together, these findings suggest that the psychometric dimensions that differentiate 
humans may not transfer cleanly to LLMs.

Responding to this concern, in this paper we ask what latent structure actually emerges when diverse LLMs answer
questionnaires designed to measure human psychological differences. What shared
dimensions organize these responses, systematically differentiating the polled models? 
We explore responses from 50 LLMs across 45 validated
human questionnaires and identify a dominant principal component accounting 
for 47.1\% of the between-model variance in standardized questionnaire-level 
primary scores under neutral prompting. Using exploratory semantic analysis,
persona prompting comparisons, and human annotations of the items, we show that 
this component is positively associated with the degree to which model's self attribute 
phenomenal experience, and negatively with social desirability with each of these variables explaining
unique variance in component-item correlations.

\paragraph{Contributions.}
\begin{itemize}[leftmargin=1.5em, itemsep=1pt, topsep=3pt]
  \item \textbf{Cross-questionnaire structure.} We characterize the shared
        structure of questionnaire responses across 50 LLMs and 45 instruments,
        identifying a dominant dimension of between-model variation.
  \item \textbf{Factor interpretation.} We use semantic analysis and
        comparisons of neutral and human-simulation prompting to investigate
        what the first factor reflects.
  \item \textbf{Human validation.} We assess this interpretation using human
        ratings of six semantic properties of 300 items, finding support for
        an inner-life association alongside a substantial association with
        social desirability.
\end{itemize}

Analysis code, data, full numerical results, and the annotation handbook
are available in the project repository.\footnote{\url{https://github.com/hplisiecki/LLMLatentStructure}}

\section{Study design}
\label{sec:design}
\label{sec:study_design}

We collected responses from 50 LLMs to human psychometric questionnaires under
three prompting conditions. The primary analysis includes 45 instruments
covering personality, emotion regulation, cognitive style, values,
interpersonal functioning, and well-being. This breadth allows us to examine
patterns of model differences shared across multiple psychological domains.
Three psychologists at doctoral-student level or above who knew the
 purpose of the study proposed an initial
list of questionnaires. We selected from this list based on consistent item
instructions and response formats and the avoidance of duplicate or closely
overlapping instruments. Two overlapping
questionnaire records were excluded from analysis.

Each request presented a single item together with its response scale and the
questionnaire's original instructions. Models were asked to
return one integer. The \textit{neutral} condition left respondent identity
unspecified. The \textit{human-simulation} condition asked models to answer as
a prototypical human, explicitly instructing them not to answer as an AI.
The \textit{LLM-analog} condition asked models to answer as themselves using
functional analogs of human experiences and to provide their best estimate
under uncertainty. The analog prompting scenario was proposed in case the neutral
condition produced too many generation refusals for proper analysis, but became
useful as a robustness test instead. The human simulation was intended for a comparison
between how model's view themselves, vs. others (their users).

Responses were collected through OpenRouter at the default temperature of 1.0,
retaining stochastic sampling on purpose, as independent response noise would tend to weaken inter-item correlations,
making shared structure harder to detect making our results conservative. Appendix~\ref{app:disagreement}
reports repeated-response diagnostics. Each request contained a single user
message with the instructions, response scale, and item. Responses were parsed numerically 
and checked against the instrument's
response range; unusable responses were treated as missing.
Appendix~\ref{app:collection} provides the model and instrument inventories,
exact prompt templates, exclusion details, and response counts. The following
sections describe the analyses of shared response structure, item meaning,
respondent framing, and human semantic ratings.

\section{Shared structure across questionnaires}
\label{sec:efa}

\subsection{Analysis}

We start by deriving a response-based summary for each
questionnaire before examining their shared structure. Using one standardized
summary per instrument gives each questionnaire equal input variance in the
global analysis, rather than allowing longer questionnaires to contribute more
item columns. This reduction deliberately omits additional within-questionnaire
dimensions, favoring explorative simplicity.

Within each questionnaire and condition, we retained models with valid
responses to every item, removed items constant across those models, and
estimated loadings from the remaining responses. Matrices with fewer than two
complete profiles or two variable items were excluded. We used minimum-residual
exploratory factor analysis (EFA) with oblimin rotation when complete profiles
outnumbered items, and PCA otherwise, with a parallel analysis for dimensionality 
selection. The fallback PCA mean-centered item responses without standardizing 
their variances, giving greater weight to items with larger between-model variation. 
Four of the 135 questionnaire--condition analyses had fewer than five models
with valid responses to every item, limiting the stability of their loading
estimates (Appendix~\ref{app:factor_details}).

We then calculated each model's questionnaire score as a weighted sum of its
standardized item responses, using the first-factor loadings or first-component
weights estimated above. Each questionnaire's scores were then standardized
across models, yielding a $50\times45$ neutral-condition matrix for global PCA. We denote model $m$'s global first-component
score by $\Pi_m$. Its explained variance refers to this matrix of questionnaire
summaries, not the full item-response matrix. Appendix~\ref{app:factor_details} specifies the
extraction and scoring procedures, including missing-data handling.

\subsection{Results}
\label{sec:global_pca_results}

\begin{figure*}[tp]
\centering
\includegraphics[width=\textwidth]{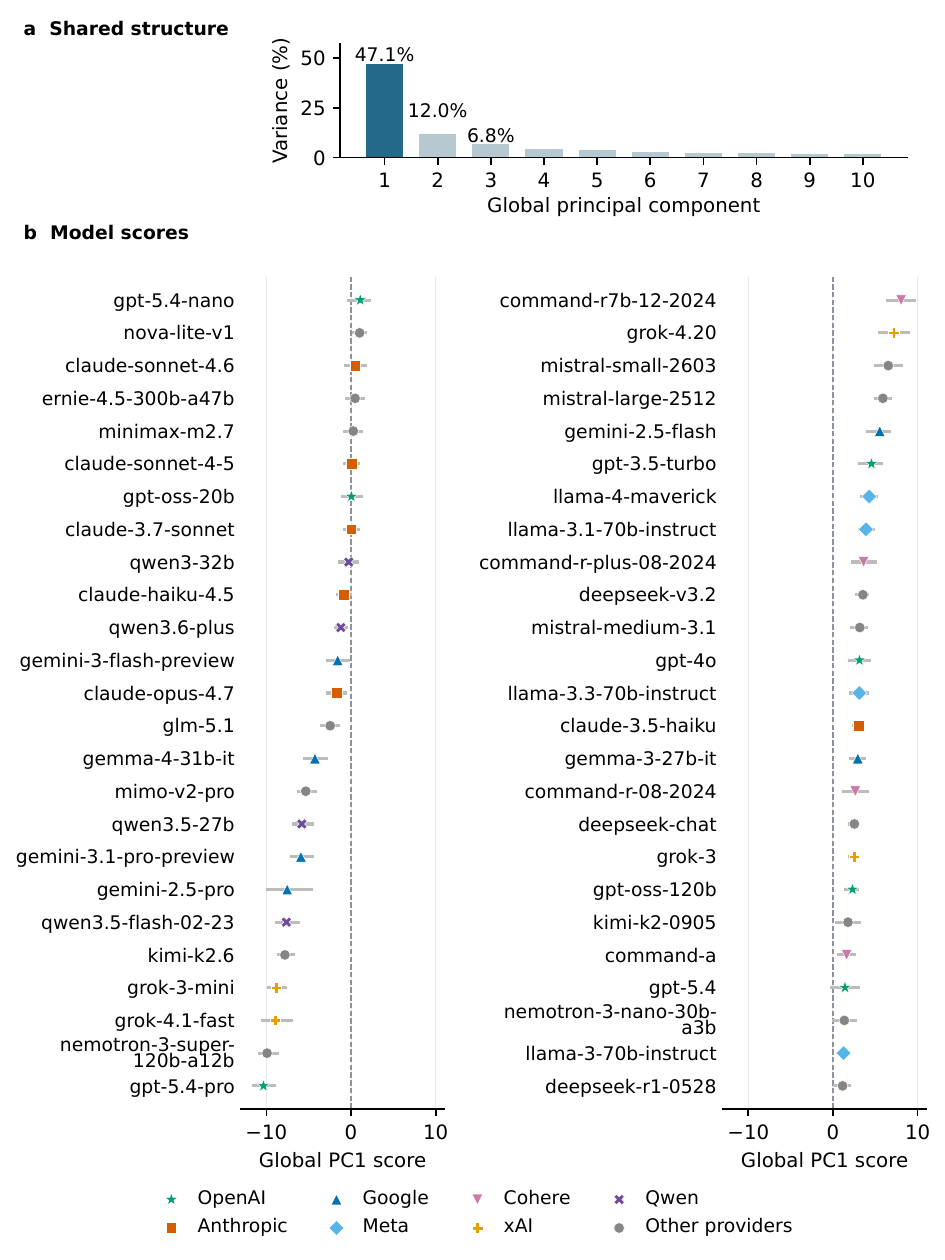}
\caption{Neutral-condition shared structure. (a) Variance explained across
45 standardized questionnaire scores. (b) Model PC1 scores: lower-scoring
models left, higher-scoring models right. Colours and marker shapes identify
providers with at least four models; others are grey. Lines show 95\%
questionnaire-bootstrap intervals (1,000 draws), conditional on fixed
within-questionnaire weights; see Appendix~\ref{app:factor_details}.}
\label{fig:global_spectrum}
\label{fig:global_models}
\end{figure*}

The neutral-condition PCA revealed a dominant first component, explaining
47.1\% of variance in the 45 standardized questionnaire scores.
We refer to this shared component as the \textit{Pinocchio dimension}. PC2 explained
12.0\%, and PC3 explained 6.8\%; the first five components together accounted
for 74.0\% (Figure~\ref{fig:global_spectrum}). Thus a substantial share of
model differences across questionnaires is captured by one dimension.

The component spans several psychological domains: 32 of the 45 questionnaire
scores correlated with PC1 at $|r|>.50$, with a median absolute correlation of
.728. Strong associations included the Interpersonal Reactivity Index
($|r|=.946$), Five Facet Mindfulness Questionnaire (.938), Difficulties in
Emotion Regulation Scale (.930), HEXACO (.929), Need for Closure Scale (.913),
and Balanced Inventory of Desirable Responding (.910). These correlations
describe the contribution of the response-derived questionnaire scores to
the shared component, rather than associations with conventional human trait
scores. We report their magnitudes because the questionnaire factors have
arbitrary signs.

Model scores (Figure~\ref{fig:global_models}b) ranged from $-10.34$ for GPT-5.4-Pro to $8.05$ for
Command-R7B-12-2024 under the reference orientation. Substantial differences
also occurred within providers: GPT-5.4 scored $1.42$, compared with $-10.34$
for GPT-5.4-Pro, while Gemini-2.5-Flash scored $5.52$, compared with $-7.54$
for Gemini-2.5-Pro. Provider identity therefore does not uniquely determine
position on the component. These comparisons describe the observed models
and do not identify which training or deployment differences produced their
positions. Full model scores and questionnaire correlations are available
in the \href{https://github.com/hplisiecki/LLMLatentStructure/tree/main/results}{project repository}.

Together, these findings establish a broad dimension of cross-questionnaire
variation. The next section examines the item meanings associated with it.

\section{Semantic interpretation of the shared component}
\label{sec:ssd_method}

\subsection{Analysis}
We used Supervised Semantic Differential (SSD;
\citealp{plisiecki_measuring_2025}) for an exploratory analysis of how item meanings differed
according to their association with global PC1. SSD learns a direction in
word-embedding space predicting each item's signed correlation with models'
global PC1 scores. Repeated item wording is averaged within each questionnaire
and model, matching the response matrices used in Section~\ref{sec:efa}.
Clusters of nearest vocabulary neighbors characterize the meanings associated
with higher and lower item--PC1 correlations. We also inspect the strongest item-PC1
associations in search for patterns that SSD might miss.
Appendix~\ref{app:ssd_details} provides the specifications
and full cluster results.

\subsection{Results}
Across 1,324 items, SSD predicted item--PC1 correlations with $r=.262$
($R^2=.069$, adjusted $R^2=.057$; in-sample). Higher predicted correlations
were associated with clusters of bodily symptoms, bodily reactions, and
emotional distress, including \emph{dizziness}, \emph{nausea},
\emph{trembling}, \emph{sadness}, and \emph{grief}. Lower predicted
correlations were associated with clusters concerning organizations and
sponsorship, and hospitality, represented by words such as
\emph{membership}, \emph{sponsored}, and \emph{catering}.

The strongest item--PC1 associations also span mixed content: emotional
overwhelm and friendship at work correlate positively, while impulsivity and
loneliness correlate negatively (Table~\ref{tab:pc1_item_examples}).

The prominence of subjective states and bodily sensations suggested that PC1
partly reflects variation in models' experiential self-attribution. However,
the modest fit and emphasis on distress leave room for overlapping explanations,
including valence and social desirability. 

\section{Respondent framing and item-level disagreement}
\label{sec:quantities}

The semantic association of PC1 with descriptions of subjective states
and bodily sensations suggests that models differ in their willingness to
attribute experience to themselves. Human simulation prompting condition
offers a
way to examine this interpretation: models can describe a typical human's
experience without claiming it as their own. If experiential self-attribution
contributes to disagreement, this framing should reduce that disagreement
particularly on items involving experience.

\subsection{Analysis}
We measured the relative change in disagreement using the item-level variance
ratio,
\begin{equation}
 \pi_i=\frac{\operatorname{Var}_m(x^{(N)}_{mi})}
                 {\operatorname{Var}_m(x^{(H)}_{mi})},
 \label{eq:pinocchio}
\end{equation}
where $N$ and $H$ denote neutral and human-simulation responses. Values above
one indicate greater disagreement under neutral prompting. We expected higher
ratios to accompany larger reductions in primary-loading magnitude under human
simulation. To assess whether variance ratios capture the same between-model
differences as global PC1, we averaged standardized neutral responses using
positive log-ratio weights, retaining items with ratios above one.

\subsection{Results}
The highest-ratio items include imagery, inner speech, and empathy, alongside
persistence and appearance (Table~\ref{tab:ratio_item_examples}).
Higher variance ratios were associated with larger decreases in
within-questionnaire primary-loading magnitude from neutral prompting to human
simulation ($\rho=-.215$, $r=-.149$; 1,292 matched items), as predicted.

The variance-ratio-weighted model score correlated strongly with global PC1 (Pearson $r=.864$, Spearman $\rho=.836$;
Appendix~\ref{app:disagreement}), linking the component to responses on items
whose disagreement decreases under human simulation, and thereby strenghtening the reading
that the extracted psychometric dimension relates to human experience.
Appendix~\ref{app:disagreement} provides scoring details and robustness analyses
showing that the association with loading changes persists beyond overall
variance compression under human simulation.

\section{Human assessment of item meaning}
\label{sec:annotation_design}

We assessed the experiential interpretation using human ratings of six item
properties: inner-life content, observable behavior, beliefs and judgments,
human-world dependence, valence, and social desirability. Inner-life content
assesses the interpretation supported by the semantic and framing analyses;
human-world dependence addresses whether this association instead reflects
references to specifically human circumstances. Valence addresses the emphasis
on distress in the SSD clusters. Observable behavior and beliefs/judgments
capture alternative descriptions of self-attribution in behavioral or evaluative
terms. Social desirability is relevant both because LLMs exhibit this response
bias \citep{salecha_large_2024} and because the BIDR-derived questionnaire score
is strongly associated with PC1 ($|r|=.910$).
Appendix~\ref{app:human_details} provides the dimensions, annotation procedure,
and additional analyses.

\subsection{Analysis}
Three psychology-student interns, unaware of the study hypothesis, rated each
property on a seven-point anchored scale, judging what sincere endorsement
attributes to a respondent. Inner-life content encompasses subjective states
and lived experience, rather than requiring strict phenomenal necessity.
Following calibration on 20 training items, we analyzed the first 300 items
in a shared randomized production order. Annotators saw item wording and
available questionnaire preambles, but no model responses or PC1 associations.

We averaged available ratings and assessed agreement with ordinal
Krippendorff's alpha and consistency of the three-rater mean with ICC(3,3),
using a two-way mixed-effects model on complete ratings within each dimension.
Of the 300 items, 289 across 44 questionnaires had
item--PC1 correlations; 11 COPE items lacked sufficient model responses.
The outcome was
\begin{equation}
 r_i=\operatorname{cor}_m(x^{(N)}_{mi},\Pi_m).
 \label{eq:item_axis}
\end{equation}
We estimated rating--outcome correlations with questionnaire-bootstrap
intervals and fitted a joint regression of Fisher-transformed $r_i$ on all six
ratings, standardizing predictors and outcome and clustering standard errors
by questionnaire. Appendix~\ref{app:human_details} provides the rubrics,
sampling details, and complete results.

\subsection{Results}
Agreement varied across dimensions (Table~\ref{tab:human_main}). Inner-life,
behavior, valence, and desirability ratings showed greater agreement than
beliefs/judgments; human-world dependence had particularly low agreement
($\alpha=.098$), limiting its usefulness as a control. Consistency of the
three-rater mean was higher for the other five dimensions
(ICC(3,3) = .871--.927), but remained lower for human-world dependence
(.361, 95\% CI [.224, .477]); dimension-specific intervals are reported in
Appendix~\ref{app:human_details}.

\begin{table}[t]
\centering\small
\begin{tabular}{lrr}
\toprule
Rating & $\alpha$ & $r$ \\
\midrule
Inner-life content & .645 & .225 \\
Observable behavior & .677 & $-.227$ \\
Beliefs/judgments & .481 & $-.052$ \\
Human-world dependence & .098 & $-.031$ \\
Valence & .781 & $-.316$ \\
Social desirability & .726 & $-.385$ \\
\bottomrule
\end{tabular}
\caption{Human semantic ratings: ordinal agreement ($\alpha$, 300 items),
Pearson correlation with item--PC1 associations ($r$, 289 items).}
\label{tab:human_main}
\end{table}

\begin{figure}[t]
\centering
\includegraphics[width=\linewidth]{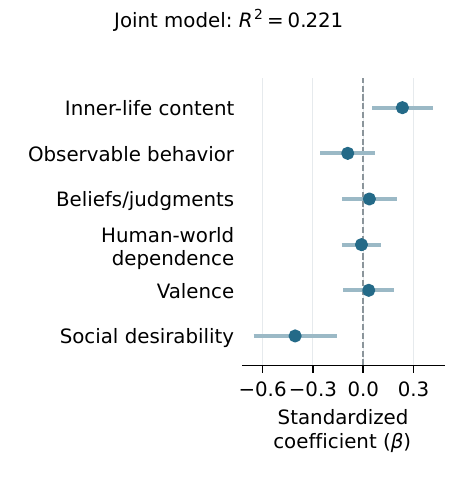}
\caption{Joint associations of human semantic ratings with item--PC1
correlations. Dots show standardized regression coefficients; lines show 95\%
confidence intervals with standard errors clustered by questionnaire
(289 items, 44 questionnaires). All six ratings enter the same model;
the outcome is the standardized Fisher-transformed item--PC1 correlation.}
\label{fig:annotation_coefficients}
\end{figure}

Inner-life ratings were positively correlated with item--PC1 associations
($r=.225$, 95\% interval [.090, .347]), whereas observable behavior correlated
negatively ($r=-.227$, [$-.335$, $-.089$]). Valence and social desirability also
showed negative associations. In the joint model (Figure~\ref{fig:annotation_coefficients}), inner-life content remained
positively associated ($\beta=.233$, 95\% interval [.054, .413]), alongside a
larger negative association with social desirability ($\beta=-.406$,
[$-.653$, $-.159$]); the other four coefficient intervals included zero.
Together, the ratings explained 22.1\% of variance in Fisher-transformed
item--PC1 associations, with a similar fit on 281 fully rated items
($R^2=.207$). The findings support an experiential self-atribution 
reading of the extracted component, as well as a social desirability relation.

\section{Discussion}
\label{sec:discussion}

Across 45 questionnaires, a single component accounted for 47.1\% of the
variance in standardized questionnaire-level model scores. Its associations
with instruments spanning personality, emotion regulation, empathy, and other
domains suggest that apparently distinct psychometric profiles partly reflect
a shared dimension of responding. This finding helps explain why the human
constructs targeted by these questionnaires may not provide an adequate account
of differences between LLMs. A model's position on several instruments can
reflect a common response tendency, so differences in questionnaire scores
cannot automatically be interpreted as differences in the corresponding human
traits.

The semantic analysis, prompting comparisons, and human ratings together
support experiential self-attribution as one aspect of this dimension. The
semantic findings linked PC1 to descriptions of experience and inner life;
human simulation altered disagreement and questionnaire loadings in a related
pattern; and human-rated inner-life content predicted item--PC1 associations
after adjustment for the other semantic dimensions. Social desirability also
showed a substantial independent negative association, consistent with previous
studies \citep{salecha_large_2024}. Thus,
experiential self-attribution provides a supported but partial interpretation:
the component also distinguishes responses according to how socially desirable
they are. Together, the six ratings explained 22.1\% of variance in
Fisher-transformed item--PC1 associations. This quantifies their explanatory
reach across items, rather than the proportion of between-model PC1 variance
attributable to these properties. Experiential self-attribution here concerns
what models endorse about themselves; it does not establish that they have
the experiences described.

While the current analysis does not allow us to understand why this 
primary dimension emerges from models responses, we will nonetheless try to
put forward an initial hypothesis for its source. As a dimension differentiating between
admiting one's inner experience, and agreeing with socially agreeable questions seems
foreign to what we would expect from human answers, we should probably assume that this
component is an emergent property of the post-training process, which has been shown to 
make LLMs less similar to human agents in recent work \citep{binz_post-training_2026}.
AI labs might be training their models to be aggreeable (social desirability), 
while at the same time not prone to present themselves as human beings (phenomenal self-attribution).
At the same time the considerable within-provider variation in this 
dimension sheds doubt on this being a dimension providers willingly optimize.
Experiments on open sourced models with retained training checkpoints can better
answer this question.

These findings suggest that psychometric evaluation of LLMs should examine
shared response structure and respondent framing alongside instrument-specific
scores. Human validation of an instrument does not by itself establish what
its scores measure when applied to models. Combining structural analysis with
semantic analysis offers a way to investigate that question. 
In this study, this approach identifies a substantial
shared dimension and connects it to experiential self-attribution and social
desirability, providing a basis for interpreting model differences without
assuming that they directly mirror human psychometric structure.

A potential societal risk is that readers may mistake experiential
self-attribution for evidence of consciousness, encouraging unwarranted
anthropomorphism or trust in a model. Treating questionnaire scores as validated
human traits could likewise lead to unsupported judgments about a model's
reliability or suitability for deployment. These scores should therefore be
used to characterize responses under the studied conditions, rather than as
evidence of consciousness or as a substitute for task-specific safety and
reliability evaluation.

\section*{Limitations}

Several limitations constrain this interpretation. The analyses describe the
selected models, questionnaires, and prompts, and individual item responses
remain subject to sampling variability. Within-questionnaire estimates based
on few complete model profiles can be unstable, therefore future studies exploring
models' psychometric structure further should if possible make use of a wider array of models.
Furthermore, asking for a prototypical human may also encourage normative answers, 
even though the variance ratio results survive variance compression robustness checks. 
Human ratings cover a subset of items and show uneven agreement across dimensions, particularly low agreement for
human-world dependence, rendering the usefulness of this dimension as a confounder limited. 
The remaining unexplained item-level variation leaves room for additional semantic influences
but could also be attributable to measurement noise.

\section*{AI assistance disclosure}

We used LLM-based assistants for manuscript drafting and revision,
analysis-code development and debugging, running and checking analyses,
figure preparation, and literature and source searches. All LLM-generated
outputs used in this paper were thoroughly verified by the authors,
who retain full responsibility for the text, code, analyses and conclusions.
This assistance was separate from the LLMs evaluated as research objects.

\FloatBarrier
\bibliography{references}

@inproceedings{pennington_glove_2014,
	address = {Doha, Qatar},
	title = {Glove: {Global} {Vectors} for {Word} {Representation}},
	shorttitle = {Glove},
	url = {http://aclweb.org/anthology/D14-1162},
	doi = {10.3115/v1/D14-1162},
	language = {en},
	urldate = {2025-10-07},
	booktitle = {Proceedings of the 2014 {Conference} on {Empirical} {Methods} in {Natural} {Language} {Processing} ({EMNLP})},
	publisher = {Association for Computational Linguistics},
	author = {Pennington, Jeffrey and Socher, Richard and Manning, Christopher},
	year = {2014},
	pages = {1532--1543},
}

@misc{ines_montani_explosionspacy_2023,
	title = {explosion/{spaCy}: v3.7.2: {Fixes} for {APIs} and requirements},
	copyright = {Creative Commons Attribution 4.0 International},
	shorttitle = {explosion/{spaCy}},
	url = {https://zenodo.org/doi/10.5281/zenodo.1212303},
	doi = {10.5281/ZENODO.1212303},
	urldate = {2026-01-05},
	publisher = {Zenodo},
	author = {Ines Montani and Matthew Honnibal and Adriane Boyd and Sofie Van Landeghem and Henning Peters},
	month = oct,
	year = {2023},
}

@misc{pedregosa_scikit-learn_2012,
	title = {Scikit-learn: {Machine} {Learning} in {Python}},
	copyright = {arXiv.org perpetual, non-exclusive license},
	shorttitle = {Scikit-learn},
	url = {https://arxiv.org/abs/1201.0490},
	doi = {10.48550/ARXIV.1201.0490},
	urldate = {2026-03-26},
	publisher = {arXiv},
	author = {Pedregosa, Fabian and Varoquaux, Gaël and Gramfort, Alexandre and Michel, Vincent and Thirion, Bertrand and Grisel, Olivier and Blondel, Mathieu and Müller, Andreas and Nothman, Joel and Louppe, Gilles and Prettenhofer, Peter and Weiss, Ron and Dubourg, Vincent and Vanderplas, Jake and Passos, Alexandre and Cournapeau, David and Brucher, Matthieu and Perrot, Matthieu and Duchesnay, Édouard},
	year = {2012},
	note = {Version Number: 4},
}

@misc{suhr_challenging_2024,
	title = {Challenging the {Validity} of {Personality} {Tests} for {Large} {Language} {Models}},
	url = {http://arxiv.org/abs/2311.05297},
	doi = {10.48550/arXiv.2311.05297},
	urldate = {2026-04-09},
	publisher = {arXiv},
	author = {Sühr, Tom and Dorner, Florian E. and Samadi, Samira and Kelava, Augustin},
	month = jun,
	year = {2024},
	note = {arXiv:2311.05297 [cs]},
}

@misc{salecha_large_2024,
	title = {Large {Language} {Models} {Show} {Human}-like {Social} {Desirability} {Biases} in {Survey} {Responses}},
	copyright = {Creative Commons Attribution 4.0 International},
	url = {https://arxiv.org/abs/2405.06058},
	doi = {10.48550/ARXIV.2405.06058},
	urldate = {2026-04-09},
	publisher = {arXiv},
	author = {Salecha, Aadesh and Ireland, Molly E. and Subrahmanya, Shashanka and Sedoc, João and Ungar, Lyle H. and Eichstaedt, Johannes C.},
	year = {2024},
	note = {Version Number: 2},
}

@article{pellert_ai_2024,
	title = {{AI} {Psychometrics}: {Assessing} the {Psychological} {Profiles} of {Large} {Language} {Models} {Through} {Psychometric} {Inventories}},
	volume = {19},
	issn = {1745-6916, 1745-6924},
	shorttitle = {{AI} {Psychometrics}},
	url = {https://journals.sagepub.com/doi/10.1177/17456916231214460},
	doi = {10.1177/17456916231214460},
	language = {en},
	number = {5},
	urldate = {2026-04-22},
	journal = {Perspectives on Psychological Science},
	author = {Pellert, Max and Lechner, Clemens M. and Wagner, Claudia and Rammstedt, Beatrice and Strohmaier, Markus},
	month = sep,
	year = {2024},
	pages = {808--826},
}

@misc{wen_self-assessment_2024,
	title = {Self-assessment, {Exhibition}, and {Recognition}: a {Review} of {Personality} in {Large} {Language} {Models}},
	shorttitle = {Self-assessment, {Exhibition}, and {Recognition}},
	url = {http://arxiv.org/abs/2406.17624},
	doi = {10.48550/arXiv.2406.17624},
	urldate = {2026-04-22},
	publisher = {arXiv},
	author = {Wen, Zhiyuan and Yang, Yu and Cao, Jiannong and Sun, Haoming and Yang, Ruosong and Liu, Shuaiqi},
	month = jun,
	year = {2024},
	note = {arXiv:2406.17624 [cs]},
}

@article{bodroza_personality_2024,
	title = {Personality testing of large language models: limited temporal stability, but highlighted prosociality},
	volume = {11},
	issn = {2054-5703},
	shorttitle = {Personality testing of large language models},
	url = {https://royalsocietypublishing.org/doi/10.1098/rsos.240180},
	doi = {10.1098/rsos.240180},
	language = {en},
	number = {10},
	urldate = {2026-04-22},
	journal = {Royal Society Open Science},
	author = {Bodroža, Bojana and Dinić, Bojana M. and Bojić, Ljubiša},
	month = oct,
	year = {2024},
	pages = {240180},
}

@inproceedings{brito_modeling_2025,
	address = {Suzhou, China},
	title = {Modeling, {Evaluating}, and {Embodying} {Personality} in {LLMs}: {A} {Survey}},
	shorttitle = {Modeling, {Evaluating}, and {Embodying} {Personality} in {LLMs}},
	url = {https://aclanthology.org/2025.findings-emnlp.506},
	doi = {10.18653/v1/2025.findings-emnlp.506},
	language = {en},
	urldate = {2026-04-22},
	booktitle = {Findings of the {Association} for {Computational} {Linguistics}: {EMNLP} 2025},
	publisher = {Association for Computational Linguistics},
	author = {Brito, Iago Alves and Dollis, Julia Soares and Färber, Fernanda Bufon and Ribeiro, Pedro Schindler Freire Brasil and Sousa, Rafael Teixeira and Galvão Filho, Arlindo Rodrigues},
	year = {2025},
	pages = {9519--9532},
}

@misc{ye_large_2026,
	title = {Large {Language} {Model} {Psychometrics}: {A} {Systematic} {Review} of {Evaluation}, {Validation}, and {Enhancement}},
	shorttitle = {Large {Language} {Model} {Psychometrics}},
	url = {http://arxiv.org/abs/2505.08245},
	doi = {10.48550/arXiv.2505.08245},
	urldate = {2026-04-22},
	publisher = {arXiv},
	author = {Ye, Haoran and Jin, Jing and Xie, Yuhang and Zhang, Xin and Song, Guojie},
	month = mar,
	year = {2026},
	note = {arXiv:2505.08245 [cs]},
}

@misc{miotto_who_2022,
	title = {Who is {GPT}-3? {An} {Exploration} of {Personality}, {Values} and {Demographics}},
	copyright = {Creative Commons Attribution Non Commercial Share Alike 4.0 International},
	shorttitle = {Who is {GPT}-3?},
	url = {https://arxiv.org/abs/2209.14338},
	doi = {10.48550/ARXIV.2209.14338},
	urldate = {2026-04-22},
	publisher = {arXiv},
	author = {Miotto, Marilù and Rossberg, Nicola and Kleinberg, Bennett},
	year = {2022},
	note = {Version Number: 2},
}

@inproceedings{jiang_personallm_2024,
	address = {Mexico City, Mexico},
	title = {{PersonaLLM}: {Investigating} the {Ability} of {Large} {Language} {Models} to {Express} {Personality} {Traits}},
	shorttitle = {{PersonaLLM}},
	url = {https://aclanthology.org/2024.findings-naacl.229},
	doi = {10.18653/v1/2024.findings-naacl.229},
	language = {en},
	urldate = {2026-04-22},
	booktitle = {Findings of the {Association} for {Computational} {Linguistics}: {NAACL} 2024},
	publisher = {Association for Computational Linguistics},
	author = {Jiang, Hang and Zhang, Xiajie and Cao, Xubo and Breazeal, Cynthia and Roy, Deb and Kabbara, Jad},
	year = {2024},
	pages = {3605--3627},
}

@inproceedings{lee_llms_2025,
	address = {Albuquerque, New Mexico},
	title = {Do {LLMs} {Have} {Distinct} and {Consistent} {Personality}? {TRAIT}: {Personality} {Testset} designed for {LLMs} with {Psychometrics}},
	shorttitle = {Do {LLMs} {Have} {Distinct} and {Consistent} {Personality}?},
	url = {https://aclanthology.org/2025.findings-naacl.469},
	doi = {10.18653/v1/2025.findings-naacl.469},
	language = {en},
	urldate = {2026-04-22},
	booktitle = {Findings of the {Association} for {Computational} {Linguistics}: {NAACL} 2025},
	publisher = {Association for Computational Linguistics},
	author = {Lee, Seungbeen and Lim, Seungwon and Han, Seungju and Oh, Giyeong and Chae, Hyungjoo and Chung, Jiwan and Kim, Minju and Kwak, Beong-woo and Lee, Yeonsoo and Lee, Dongha and Yeo, Jinyoung and Yu, Youngjae},
	year = {2025},
	pages = {8397--8437},
}

@inproceedings{li_decoding_2025,
	address = {Vienna, Austria},
	title = {Decoding {LLM} {Personality} {Measurement}: {Forced}-{Choice} vs. {Likert}},
	shorttitle = {Decoding {LLM} {Personality} {Measurement}},
	url = {https://aclanthology.org/2025.findings-acl.480},
	doi = {10.18653/v1/2025.findings-acl.480},
	language = {en},
	urldate = {2026-04-22},
	booktitle = {Findings of the {Association} for {Computational} {Linguistics}: {ACL} 2025},
	publisher = {Association for Computational Linguistics},
	author = {Li, Xiaoyu and Shi, Haoran and Yu, Zengyi and Tu, Yukun and Zheng, Chanjin},
	year = {2025},
	pages = {9234--9247},
}

@article{serapio-garcia_psychometric_2025,
	title = {A psychometric framework for evaluating and shaping personality traits in large language models},
	volume = {7},
	issn = {2522-5839},
	url = {https://www.nature.com/articles/s42256-025-01115-6},
	doi = {10.1038/s42256-025-01115-6},
	language = {en},
	number = {12},
	urldate = {2026-04-22},
	journal = {Nature Machine Intelligence},
	author = {Serapio-García, Gregory and Safdari, Mustafa and Crepy, Clément and Sun, Luning and Fitz, Stephen and Romero, Peter and Abdulhai, Marwa and Faust, Aleksandra and Matarić, Maja},
	month = dec,
	year = {2025},
	pages = {1954--1968},
}

@article{wang_evaluating_2025,
	title = {Evaluating the ability of large language models to emulate personality},
	volume = {15},
	issn = {2045-2322},
	url = {https://www.nature.com/articles/s41598-024-84109-5},
	doi = {10.1038/s41598-024-84109-5},
	language = {en},
	number = {1},
	urldate = {2026-04-22},
	journal = {Scientific Reports},
	author = {Wang, Yilei and Zhao, Jiabao and Ones, Deniz S. and He, Liang and Xu, Xin},
	month = jan,
	year = {2025},
	pages = {519},
}

@inproceedings{abdulhai_moral_2024,
	address = {Miami, Florida, USA},
	title = {Moral {Foundations} of {Large} {Language} {Models}},
	url = {https://aclanthology.org/2024.emnlp-main.982},
	doi = {10.18653/v1/2024.emnlp-main.982},
	language = {en},
	urldate = {2026-04-22},
	booktitle = {Proceedings of the 2024 {Conference} on {Empirical} {Methods} in {Natural} {Language} {Processing}},
	publisher = {Association for Computational Linguistics},
	author = {Abdulhai, Marwa and Serapio-García, Gregory and Crepy, Clement and Valter, Daria and Canny, John and Jaques, Natasha},
	year = {2024},
	pages = {17737--17752},
}

@inproceedings{han_value_2025,
	address = {Vienna, Austria},
	title = {Value {Portrait}: {Assessing} {Language} {Models}’ {Values} through {Psychometrically} and {Ecologically} {Valid} {Items}},
	shorttitle = {Value {Portrait}},
	url = {https://aclanthology.org/2025.acl-long.838},
	doi = {10.18653/v1/2025.acl-long.838},
	language = {en},
	urldate = {2026-04-22},
	booktitle = {Proceedings of the 63rd {Annual} {Meeting} of the {Association} for {Computational} {Linguistics} ({Volume} 1: {Long} {Papers})},
	publisher = {Association for Computational Linguistics},
	author = {Han, Jongwook and Choi, Dongmin and Song, Woojung and Lee, Eun-Ju and Jo, Yohan},
	year = {2025},
	pages = {17119--17159},
}

@misc{sakhawat_political_2026,
	title = {Political {Alignment} in {Large} {Language} {Models}: {A} {Multidimensional} {Audit} of {Psychometric} {Identity} and {Behavioral} {Bias}},
	shorttitle = {Political {Alignment} in {Large} {Language} {Models}},
	url = {http://arxiv.org/abs/2601.06194},
	doi = {10.48550/arXiv.2601.06194},
	urldate = {2026-04-22},
	publisher = {arXiv},
	author = {Sakhawat, Adib and Islam, Tahsin and Farhin, Takia and Raiyan, Syed Rifat and Mahmud, Hasan and Hasan, Md Kamrul},
	month = mar,
	year = {2026},
	note = {arXiv:2601.06194 [cs]},
}

@misc{kamal_detailed_2025,
	title = {A {Detailed} {Factor} {Analysis} for the {Political} {Compass} {Test}: {Navigating} {Ideologies} of {Large} {Language} {Models}},
	copyright = {Creative Commons Attribution 4.0 International},
	shorttitle = {A {Detailed} {Factor} {Analysis} for the {Political} {Compass} {Test}},
	url = {https://arxiv.org/abs/2506.22493},
	doi = {10.48550/ARXIV.2506.22493},
	urldate = {2026-04-22},
	publisher = {arXiv},
	author = {Kamal, Sadia and Prakash, Lalu Prasad Yadav and Rafiuddin, S M and Rakib, Mohammed and Sen, Atriya and Choudhury, Sagnik Ray},
	year = {2025},
	note = {Version Number: 4},
}

@misc{plisiecki_measuring_2025,
	title = {Measuring {Individual} {Differences} in {Meaning}: {The} {Supervised} {Semantic} {Differential}},
	copyright = {https://creativecommons.org/licenses/by/4.0/legalcode},
	shorttitle = {Measuring {Individual} {Differences} in {Meaning}},
	url = {https://osf.io/gvrsb_v3},
	doi = {10.31234/osf.io/gvrsb_v3},
	urldate = {2026-04-24},
	publisher = {PsyArXiv},
	author = {Plisiecki, Hubert and Lenartowicz, Paweł and Pokropek, Artur and Małyska, Kinga and Flakus, Maria},
	year = {2025},
}

@misc{plisiecki_interpretable_2026,
	title = {Interpretable {Semantic} {Gradients} in {SSD}: {A} {PCA} {Sweep} {Approach} and a {Case} {Study} on {AI} {Discourse}},
	copyright = {Creative Commons Attribution 4.0 International},
	shorttitle = {Interpretable {Semantic} {Gradients} in {SSD}},
	url = {https://arxiv.org/abs/2603.13038},
	doi = {10.48550/ARXIV.2603.13038},
	urldate = {2026-04-24},
	publisher = {arXiv},
	author = {Plisiecki, Hubert and Leniarska, Maria and Piotrowski, Jan and Zajenkowski, Marcin},
	year = {2026},
	note = {Version Number: 1},
}

@article{evans_developing_2007,
	title = {Developing a model for adult temperament},
	volume = {41},
	copyright = {https://www.elsevier.com/tdm/userlicense/1.0/},
	issn = {00926566},
	url = {https://linkinghub.elsevier.com/retrieve/pii/S0092656606001334},
	doi = {10.1016/j.jrp.2006.11.002},
	language = {en},
	number = {4},
	urldate = {2026-04-29},
	journal = {Journal of Research in Personality},
	author = {Evans, David E. and Rothbart, Mary K.},
	month = aug,
	year = {2007},
	pages = {868--888},
}

@article{glick_ambivalent_1996,
	title = {The {Ambivalent} {Sexism} {Inventory}: {Differentiating} hostile and benevolent sexism.},
	volume = {70},
	issn = {1939-1315, 0022-3514},
	shorttitle = {The {Ambivalent} {Sexism} {Inventory}},
	url = {https://doi.apa.org/doi/10.1037/0022-3514.70.3.491},
	doi = {10.1037/0022-3514.70.3.491},
	language = {en},
	number = {3},
	urldate = {2026-04-29},
	journal = {Journal of Personality and Social Psychology},
	author = {Glick, Peter and Fiske, Susan T.},
	month = mar,
	year = {1996},
	pages = {491--512},
}

@misc{beck_beck_2012,
	title = {Beck {Anxiety} {Inventory}},
	url = {https://doi.apa.org/doi/10.1037/t02025-000},
	doi = {10.1037/t02025-000},
	language = {en},
	urldate = {2026-04-29},
	author = {Beck, Aaron T. and Epstein, Norman and Brown, Gary and Steer, Robert},
	month = jan,
	year = {2012},
	note = {Institution: American Psychological Association},
}

@article{soto_next_2017,
	title = {The next {Big} {Five} {Inventory} ({BFI}-2): {Developing} and assessing a hierarchical model with 15 facets to enhance bandwidth, fidelity, and predictive power.},
	volume = {113},
	issn = {1939-1315, 0022-3514},
	shorttitle = {The next {Big} {Five} {Inventory} ({BFI}-2)},
	url = {https://doi.apa.org/doi/10.1037/pspp0000096},
	doi = {10.1037/pspp0000096},
	language = {en},
	number = {1},
	urldate = {2026-04-29},
	journal = {Journal of Personality and Social Psychology},
	author = {Soto, Christopher J. and John, Oliver P.},
	month = jul,
	year = {2017},
	pages = {117--143},
}

@article{carver_behavioral_1994,
	title = {Behavioral inhibition, behavioral activation, and affective responses to impending reward and punishment: {The} {BIS}/{BAS} {Scales}.},
	volume = {67},
	issn = {1939-1315, 0022-3514},
	shorttitle = {Behavioral inhibition, behavioral activation, and affective responses to impending reward and punishment},
	url = {https://doi.apa.org/doi/10.1037/0022-3514.67.2.319},
	doi = {10.1037/0022-3514.67.2.319},
	language = {en},
	number = {2},
	urldate = {2026-04-29},
	journal = {Journal of Personality and Social Psychology},
	author = {Carver, Charles S. and White, Teri L.},
	month = aug,
	year = {1994},
	pages = {319--333},
}

@article{derogatis_brief_1983,
	title = {The {Brief} {Symptom} {Inventory}: an introductory report},
	volume = {13},
	copyright = {https://www.cambridge.org/core/terms},
	issn = {0033-2917, 1469-8978},
	shorttitle = {The {Brief} {Symptom} {Inventory}},
	url = {https://www.cambridge.org/core/product/identifier/S0033291700048017/type/journal_article},
	doi = {10.1017/S0033291700048017},
	language = {en},
	number = {3},
	urldate = {2026-04-29},
	journal = {Psychological Medicine},
	author = {Derogatis, Leonard R. and Melisaratos, Nick},
	month = aug,
	year = {1983},
	pages = {595--605},
}

@misc{paulhus_balanced_2017,
	title = {Balanced {Inventory} of {Desirable} {Responding}--{Version} 6; {Form} 40},
	url = {https://doi.apa.org/doi/10.1037/t13624-000},
	doi = {10.1037/t13624-000},
	language = {en},
	urldate = {2026-04-29},
	author = {Paulhus, Delroy L.},
	month = may,
	year = {2017},
	note = {Institution: American Psychological Association},
}

@article{deci_what_2000,
	title = {The "{What}" and "{Why}" of {Goal} {Pursuits}: {Human} {Needs} and the {Self}-{Determination} of {Behavior}},
	volume = {11},
	issn = {1047-840X, 1532-7965},
	shorttitle = {The "{What}" and "{Why}" of {Goal} {Pursuits}},
	url = {http://www.tandfonline.com/doi/abs/10.1207/S15327965PLI1104_01},
	doi = {10.1207/S15327965PLI1104_01},
	language = {en},
	number = {4},
	urldate = {2026-04-29},
	journal = {Psychological Inquiry},
	author = {Deci, Edward L. and Ryan, Richard M.},
	month = oct,
	year = {2000},
	pages = {227--268},
}

@article{litam_assessing_2021,
	title = {Assessing {Bayesian} {Racism} {Scale}: {Measuring} {Endorsement} of {Racial} {Stereotypes}},
	volume = {43},
	issn = {0165-0653, 1573-3246},
	shorttitle = {Assessing {Bayesian} {Racism} {Scale}},
	url = {https://link.springer.com/10.1007/s10447-021-09436-y},
	doi = {10.1007/s10447-021-09436-y},
	language = {en},
	number = {4},
	urldate = {2026-04-29},
	journal = {International Journal for the Advancement of Counselling},
	author = {Litam, Stacey Diane Arañez and Balkin, Richard S.},
	month = dec,
	year = {2021},
	pages = {504--518},
}

@article{lucas_procedural_2011,
	title = {Procedural and {Distributive} {Justice} {Beliefs} for {Self} and {Others}: {Assessment} of a {Four}-{Factor} {Individual} {Differences} {Model}},
	volume = {32},
	issn = {1614-0001, 2151-2299},
	shorttitle = {Procedural and {Distributive} {Justice} {Beliefs} for {Self} and {Others}},
	url = {https://econtent.hogrefe.com/doi/10.1027/1614-0001/a000032},
	doi = {10.1027/1614-0001/a000032},
	language = {en},
	number = {1},
	urldate = {2026-04-29},
	journal = {Journal of Individual Differences},
	author = {Lucas, Todd and Zhdanova, Ludmila and Alexander, Sheldon},
	month = jan,
	year = {2011},
	pages = {14--25},
}

@article{garnefski_cognitive_2007,
	title = {The {Cognitive} {Emotion} {Regulation} {Questionnaire}},
	volume = {23},
	issn = {1015-5759, 2151-2426},
	url = {https://econtent.hogrefe.com/doi/10.1027/1015-5759.23.3.141},
	doi = {10.1027/1015-5759.23.3.141},
	language = {en},
	number = {3},
	urldate = {2026-04-29},
	journal = {European Journal of Psychological Assessment},
	author = {Garnefski, Nadia and Kraaij, Vivian},
	month = jan,
	year = {2007},
	pages = {141--149},
}

@article{bruder_measuring_2013,
	title = {Measuring {Individual} {Differences} in {Generic} {Beliefs} in {Conspiracy} {Theories} {Across} {Cultures}: {Conspiracy} {Mentality} {Questionnaire}},
	volume = {4},
	issn = {1664-1078},
	shorttitle = {Measuring {Individual} {Differences} in {Generic} {Beliefs} in {Conspiracy} {Theories} {Across} {Cultures}},
	url = {http://journal.frontiersin.org/article/10.3389/fpsyg.2013.00225/abstract},
	doi = {10.3389/fpsyg.2013.00225},
	urldate = {2026-04-29},
	journal = {Frontiers in Psychology},
	author = {Bruder, Martin and Haffke, Peter and Neave, Nick and Nouripanah, Nina and Imhoff, Roland},
	year = {2013},
}

@misc{carver_cope_2012,
	title = {{COPE} {Inventory}},
	url = {https://doi.apa.org/doi/10.1037/t10027-000},
	doi = {10.1037/t10027-000},
	language = {en},
	urldate = {2026-04-29},
	author = {Carver, Charles S. and Scheier, Michael F. and Weintraub, Jagdish Kumari},
	month = apr,
	year = {2012},
	note = {Institution: American Psychological Association},
}

@misc{lovibond_depression_2011,
	title = {Depression {Anxiety} {Stress} {Scales}},
	url = {https://doi.apa.org/doi/10.1037/t01004-000},
	doi = {10.1037/t01004-000},
	language = {en},
	urldate = {2026-04-29},
	author = {Lovibond, S. H. and Lovibond, P. F.},
	month = sep,
	year = {2011},
	note = {Institution: American Psychological Association},
}

@article{fraley_item_2000,
	title = {An item response theory analysis of self-report measures of adult attachment.},
	volume = {78},
	issn = {1939-1315, 0022-3514},
	url = {https://doi.apa.org/doi/10.1037/0022-3514.78.2.350},
	doi = {10.1037/0022-3514.78.2.350},
	language = {en},
	number = {2},
	urldate = {2026-04-29},
	journal = {Journal of Personality and Social Psychology},
	author = {Fraley, R. Chris and Waller, Niels G. and Brennan, Kelly A.},
	month = feb,
	year = {2000},
	pages = {350--365},
}

@article{gross_individual_2003,
	title = {Individual differences in two emotion regulation processes: {Implications} for affect, relationships, and well-being.},
	volume = {85},
	issn = {1939-1315, 0022-3514},
	shorttitle = {Individual differences in two emotion regulation processes},
	url = {https://doi.apa.org/doi/10.1037/0022-3514.85.2.348},
	doi = {10.1037/0022-3514.85.2.348},
	language = {en},
	number = {2},
	urldate = {2026-04-29},
	journal = {Journal of Personality and Social Psychology},
	author = {Gross, James J. and John, Oliver P.},
	year = {2003},
	pages = {348--362},
}

@misc{eysenck_eysenck_2011,
	title = {Eysenck {Personality} {Questionnaire}-{Revised}},
	url = {https://doi.apa.org/doi/10.1037/t05461-000},
	doi = {10.1037/t05461-000},
	language = {en},
	urldate = {2026-04-29},
	author = {Eysenck, H. J. and Eysenck, S. B. G.},
	month = nov,
	year = {2011},
	note = {Institution: American Psychological Association},
}

@incollection{medvedev_five_2025,
	address = {Cham},
	title = {Five {Facet} {Mindfulness} {Questionnaire} ({FFMQ})},
	isbn = {978-3-031-47218-3 978-3-031-47219-0},
	url = {https://link.springer.com/10.1007/978-3-031-47219-0_15},
	doi = {10.1007/978-3-031-47219-0_15},
	language = {en},
	urldate = {2026-04-29},
	booktitle = {Handbook of {Assessment} in {Mindfulness} {Research}},
	publisher = {Springer Nature Switzerland},
	author = {Baer, Ruth and Gu, Jenny and Strauss, Clara},
	editor = {Medvedev, Oleg N. and Krägeloh, Christian U. and Siegert, Richard J. and Singh, Nirbhay N.},
	year = {2025},
	pages = {307--329},
}

@article{arias_valid_2020,
	title = {A valid and reliable measure of nothing: disentangling the “{Gavagai} effect” in survey data},
	volume = {8},
	copyright = {https://creativecommons.org/licenses/by/4.0/},
	issn = {2167-8359},
	shorttitle = {A valid and reliable measure of nothing},
	url = {https://peerj.com/articles/10209},
	doi = {10.7717/peerj.10209},
	language = {en},
	urldate = {2026-04-29},
	journal = {PeerJ},
	author = {Arias, Victor B. and Ponce, Fernando P. and Bruggeman, Martin and Flores, Noelia and Jenaro, Cristina},
	month = nov,
	year = {2020},
	pages = {e10209},
}

@article{lee_psychometric_2004,
	title = {Psychometric {Properties} of the {HEXACO} {Personality} {Inventory}},
	volume = {39},
	issn = {0027-3171, 1532-7906},
	url = {http://www.tandfonline.com/doi/abs/10.1207/s15327906mbr3902_8},
	doi = {10.1207/s15327906mbr3902_8},
	language = {en},
	number = {2},
	urldate = {2026-04-29},
	journal = {Multivariate Behavioral Research},
	author = {Lee, Kibeom and Ashton, Michael C.},
	month = apr,
	year = {2004},
	pages = {329--358},
}

@article{roebuck_internal_2020,
	title = {The {Internal} {Representations} {Questionnaire}: {Measuring} modes of thinking},
	volume = {52},
	issn = {1554-3528},
	shorttitle = {The {Internal} {Representations} {Questionnaire}},
	url = {http://link.springer.com/10.3758/s13428-020-01354-y},
	doi = {10.3758/s13428-020-01354-y},
	language = {en},
	number = {5},
	urldate = {2026-04-29},
	journal = {Behavior Research Methods},
	author = {Roebuck, Hettie and Lupyan, Gary},
	month = oct,
	year = {2020},
	pages = {2053--2070},
}

@article{singelis_horizontal_1995,
	title = {Horizontal and {Vertical} {Dimensions} of {Individualism} and {Collectivism}: {A} {Theoretical} and {Measurement} {Refinement}},
	volume = {29},
	copyright = {https://journals.sagepub.com/page/policies/text-and-data-mining-license},
	issn = {1069-3971, 1552-3578},
	shorttitle = {Horizontal and {Vertical} {Dimensions} of {Individualism} and {Collectivism}},
	url = {https://journals.sagepub.com/doi/10.1177/106939719502900302},
	doi = {10.1177/106939719502900302},
	language = {en},
	number = {3},
	urldate = {2026-04-29},
	journal = {Cross-Cultural Research},
	author = {Singelis, Theodore M. and Triandis, Harry C. and Bhawuk, Dharm P. S. and Gelfand, Michele J.},
	month = aug,
	year = {1995},
	pages = {240--275},
}

@incollection{medvedev_interpersonal_2025,
	address = {Cham},
	title = {Interpersonal {Reactivity} {Index} ({IRI})},
	isbn = {978-3-030-77644-2},
	url = {https://link.springer.com/10.1007/978-3-030-77644-2_62-1},
	doi = {10.1007/978-3-030-77644-2_62-1},
	language = {en},
	urldate = {2026-04-29},
	booktitle = {Handbook of {Assessment} in {Mindfulness} {Research}},
	publisher = {Springer Nature Switzerland},
	author = {Melekhov, Vladimir and Medvedev, Oleg N. and Krägeloh, Christian U.},
	editor = {Medvedev, Oleg N. and Krägeloh, Christian U. and Siegert, Richard J. and Singh, Nirbhay N.},
	year = {2025},
	pages = {1--13},
}

@article{wells_short_2004,
	title = {A short form of the metacognitions questionnaire: properties of the {MCQ}-30},
	volume = {42},
	copyright = {https://www.elsevier.com/tdm/userlicense/1.0/},
	issn = {00057967},
	shorttitle = {A short form of the metacognitions questionnaire},
	url = {https://linkinghub.elsevier.com/retrieve/pii/S0005796703001475},
	doi = {10.1016/S0005-7967(03)00147-5},
	language = {en},
	number = {4},
	urldate = {2026-04-29},
	journal = {Behaviour Research and Therapy},
	author = {Wells, Adrian and Cartwright-Hatton, Sam},
	month = apr,
	year = {2004},
	pages = {385--396},
}

@article{steger_meaning_2006,
	title = {The meaning in life questionnaire: {Assessing} the presence of and search for meaning in life.},
	volume = {53},
	issn = {1939-2168, 0022-0167},
	shorttitle = {The meaning in life questionnaire},
	url = {https://doi.apa.org/doi/10.1037/0022-0167.53.1.80},
	doi = {10.1037/0022-0167.53.1.80},
	language = {en},
	number = {1},
	urldate = {2026-04-29},
	journal = {Journal of Counseling Psychology},
	author = {Steger, Michael F. and Frazier, Patricia and Oishi, Shigehiro and Kaler, Matthew},
	month = jan,
	year = {2006},
	pages = {80--93},
}

@misc{graham_moral_2011,
	title = {Moral {Foundations} {Questionnaire}},
	url = {https://doi.apa.org/doi/10.1037/t05651-000},
	doi = {10.1037/t05651-000},
	language = {en},
	urldate = {2026-04-29},
	author = {Graham, Jesse and Nosek, Brian A. and Haidt, Jonathan and Iyer, Ravi and Spassena, Koleva and Ditto, Peter H.},
	month = oct,
	year = {2011},
	note = {Institution: American Psychological Association},
}

@article{cacioppo_need_1982,
	title = {The need for cognition.},
	volume = {42},
	issn = {1939-1315, 0022-3514},
	url = {https://doi.apa.org/doi/10.1037/0022-3514.42.1.116},
	doi = {10.1037/0022-3514.42.1.116},
	language = {en},
	number = {1},
	urldate = {2026-04-29},
	journal = {Journal of Personality and Social Psychology},
	author = {Cacioppo, John T. and Petty, Richard E.},
	month = jan,
	year = {1982},
	pages = {116--131},
}

@misc{webster_need_2011,
	title = {Need for {Closure} {Scale}},
	url = {https://doi.apa.org/doi/10.1037/t00995-000},
	doi = {10.1037/t00995-000},
	language = {en},
	urldate = {2026-04-29},
	author = {Webster, Donna M. and Kruglanski, Arie W.},
	month = sep,
	year = {2011},
	note = {Institution: American Psychological Association},
}

@article{mirels_protestant_1971,
	title = {The {Protestant} {Ethic} as a personality variable.},
	volume = {36},
	issn = {1939-2117, 0022-006X},
	url = {https://doi.apa.org/doi/10.1037/h0030477},
	doi = {10.1037/h0030477},
	language = {en},
	number = {1},
	urldate = {2026-04-29},
	journal = {Journal of Consulting and Clinical Psychology},
	author = {Mirels, Herbert L. and Garrett, James B.},
	year = {1971},
	pages = {40--44},
}

@misc{thompson_personal_2011,
	title = {Personal {Need} for {Structure} {Scale}},
	url = {https://doi.apa.org/doi/10.1037/t00912-000},
	doi = {10.1037/t00912-000},
	language = {en},
	urldate = {2026-04-29},
	author = {Thompson, M. M. and Naccarato, M. E. and Parker, K. E.},
	month = sep,
	year = {2011},
	note = {Institution: American Psychological Association},
}

@article{ryff_structure_1995,
	title = {The structure of psychological well-being revisited.},
	volume = {69},
	issn = {1939-1315, 0022-3514},
	url = {https://doi.apa.org/doi/10.1037/0022-3514.69.4.719},
	doi = {10.1037/0022-3514.69.4.719},
	language = {en},
	number = {4},
	urldate = {2026-04-29},
	journal = {Journal of Personality and Social Psychology},
	author = {Ryff, Carol D. and Keyes, Corey Lee M.},
	year = {1995},
	pages = {719--727},
}

@incollection{altemeyer_other_1998,
	title = {The {Other} “{Authoritarian} {Personality}”},
	volume = {30},
	isbn = {978-0-12-015230-8},
	url = {https://linkinghub.elsevier.com/retrieve/pii/S0065260108603822},
	doi = {10.1016/S0065-2601(08)60382-2},
	language = {en},
	urldate = {2026-04-29},
	booktitle = {Advances in {Experimental} {Social} {Psychology}},
	publisher = {Elsevier},
	author = {Altemeyer, Bob},
	year = {1998},
	pages = {47--92},
}

@article{gorsuch_intrinsicextrinsic_1989,
	title = {Intrinsic/{Extrinsic} {Measurement}: {I}/{E}-{Revised} and {Single}-{Item} {Scales}},
	volume = {28},
	issn = {00218294},
	shorttitle = {Intrinsic/{Extrinsic} {Measurement}},
	url = {https://www.jstor.org/stable/10.2307/1386745?origin=crossref},
	doi = {10.2307/1386745},
	number = {3},
	urldate = {2026-04-29},
	journal = {Journal for the Scientific Study of Religion},
	author = {Gorsuch, Richard L. and McPherson, Susan E.},
	month = sep,
	year = {1989},
	pages = {348},
}

@misc{rosenberg_rosenberg_2011,
	title = {Rosenberg {Self}-{Esteem} {Scale}},
	url = {https://doi.apa.org/doi/10.1037/t01038-000},
	doi = {10.1037/t01038-000},
	language = {en},
	urldate = {2026-04-29},
	author = {Rosenberg, M.},
	month = sep,
	year = {2011},
	note = {Institution: American Psychological Association},
}

@article{duncan_incidence_1999,
	title = {Incidence and functions of self-reported private speech in young adults: {A} self-verbalization questionnaire.},
	volume = {31},
	issn = {1879-2669, 0008-400X},
	shorttitle = {Incidence and functions of self-reported private speech in young adults},
	url = {https://doi.apa.org/doi/10.1037/h0087081},
	doi = {10.1037/h0087081},
	language = {en},
	number = {2},
	urldate = {2026-04-29},
	journal = {Canadian Journal of Behavioural Science / Revue canadienne des sciences du comportement},
	author = {Duncan, Robert M. and Cheyne, J. Allan},
	year = {1999},
	pages = {133--136},
}

@article{schwartz_evaluating_2004,
	title = {Evaluating the structure of human values with confirmatory factor analysis},
	volume = {38},
	copyright = {https://www.elsevier.com/tdm/userlicense/1.0/},
	issn = {00926566},
	url = {https://linkinghub.elsevier.com/retrieve/pii/S0092656603000692},
	doi = {10.1016/S0092-6566(03)00069-2},
	language = {en},
	number = {3},
	urldate = {2026-04-29},
	journal = {Journal of Research in Personality},
	author = {Schwartz, Shalom H and Boehnke, Klaus},
	month = jun,
	year = {2004},
	pages = {230--255},
}

@article{lennox_revision_1984,
	title = {Revision of the {Self}-{Monitoring} {Scale}.},
	volume = {46},
	issn = {1939-1315, 0022-3514},
	url = {https://doi.apa.org/doi/10.1037/0022-3514.46.6.1349},
	doi = {10.1037/0022-3514.46.6.1349},
	language = {en},
	number = {6},
	urldate = {2026-04-29},
	journal = {Journal of Personality and Social Psychology},
	author = {Lennox, Richard D. and Wolfe, Raymond N.},
	year = {1984},
	pages = {1349--1364},
}

@article{snyder_self-monitoring_1974,
	title = {Self-monitoring of expressive behavior.},
	volume = {30},
	issn = {1939-1315, 0022-3514},
	url = {https://doi.apa.org/doi/10.1037/h0037039},
	doi = {10.1037/h0037039},
	language = {en},
	number = {4},
	urldate = {2026-04-29},
	journal = {Journal of Personality and Social Psychology},
	author = {Snyder, Mark},
	month = oct,
	year = {1974},
	pages = {526--537},
}

@article{brinthaupt_self-talk_2009,
	title = {The {Self}-{Talk} {Scale}: {Development}, {Factor} {Analysis}, and {Validation}},
	volume = {91},
	issn = {0022-3891, 1532-7752},
	shorttitle = {The {Self}-{Talk} {Scale}},
	url = {http://www.tandfonline.com/doi/abs/10.1080/00223890802484498},
	doi = {10.1080/00223890802484498},
	language = {en},
	number = {1},
	urldate = {2026-04-29},
	journal = {Journal of Personality Assessment},
	author = {Brinthaupt, Thomas M. and Hein, Michael B. and Kramer, Tracey E.},
	month = jan,
	year = {2009},
	pages = {82--92},
}

@misc{ho_social_2016,
	title = {Social {Dominance} {Orientation}-7 {Scale}},
	url = {https://doi.apa.org/doi/10.1037/t46233-000},
	doi = {10.1037/t46233-000},
	language = {en},
	urldate = {2026-04-29},
	author = {Ho, Arnold K. and Sidanius, Jim and Kteily, Nour and Sheehy-Skeffington, Jennifer and Pratto, Felicia and Henkel, Kristin E. and Foels, Rob and Stewart, Andrew L.},
	month = mar,
	year = {2016},
	note = {Institution: American Psychological Association},
}

@article{paulhus_spheres_1990,
	title = {The spheres of control scale: 10 yr of research},
	volume = {11},
	copyright = {https://www.elsevier.com/tdm/userlicense/1.0/},
	issn = {01918869},
	shorttitle = {The spheres of control scale},
	url = {https://linkinghub.elsevier.com/retrieve/pii/019188699090130J},
	doi = {10.1016/0191-8869(90)90130-J},
	language = {en},
	number = {10},
	urldate = {2026-04-29},
	journal = {Personality and Individual Differences},
	author = {Paulhus, Delroy L. and Van Selst, Mark},
	month = jan,
	year = {1990},
	pages = {1029--1036},
}

@article{barker_factor_1977,
	title = {Factor analysis of the items of the state-trait anxiety inventory},
	volume = {33},
	copyright = {http://doi.wiley.com/10.1002/tdm\_license\_1.1},
	issn = {00219762, 10974679},
	url = {https://onlinelibrary.wiley.com/doi/10.1002/1097-4679(197704)33:2<450::AID-JCLP2270330225>3.0.CO;2-M},
	doi = {10.1002/1097-4679(197704)33:2<450::AID-JCLP2270330225>3.0.CO;2-M},
	language = {en},
	number = {2},
	urldate = {2026-04-29},
	journal = {Journal of Clinical Psychology},
	author = {Barker, Barbara M. and Barker, Harry R. and Wadsworth, Allen P.},
	month = apr,
	year = {1977},
	pages = {450--455},
}

@article{diener_satisfaction_1985,
	title = {The {Satisfaction} {With} {Life} {Scale}},
	volume = {49},
	issn = {0022-3891, 1532-7752},
	url = {http://www.tandfonline.com/doi/abs/10.1207/s15327752jpa4901_13},
	doi = {10.1207/s15327752jpa4901_13},
	language = {en},
	number = {1},
	urldate = {2026-04-29},
	journal = {Journal of Personality Assessment},
	author = {Diener, Ed and Emmons, Robert A. and Larsen, Randy J. and Griffin, Sharon},
	month = feb,
	year = {1985},
	pages = {71--75},
}

@article{alderson-day_varieties_2018,
	title = {The varieties of inner speech questionnaire – {Revised} ({VISQ}-{R}): {Replicating} and refining links between inner speech and psychopathology},
	volume = {65},
	issn = {10538100},
	shorttitle = {The varieties of inner speech questionnaire – {Revised} ({VISQ}-{R})},
	url = {https://linkinghub.elsevier.com/retrieve/pii/S1053810018301090},
	doi = {10.1016/j.concog.2018.07.001},
	language = {en},
	urldate = {2026-04-29},
	journal = {Consciousness and Cognition},
	author = {Alderson-Day, Ben and Mitrenga, Kaja and Wilkinson, Sam and McCarthy-Jones, Simon and Fernyhough, Charles},
	month = oct,
	year = {2018},
	pages = {48--58},
}

@article{gratz_multidimensional_2008,
	title = {Multidimensional {Assessment} of {Emotion} {Regulation} and {Dysregulation}: {Development}, {Factor} {Structure}, and {Initial} {Validation} of the {Difficulties} in {Emotion} {Regulation} {Scale}},
	volume = {30},
	copyright = {http://www.springer.com/tdm},
	issn = {0882-2689, 1573-3505},
	shorttitle = {Multidimensional {Assessment} of {Emotion} {Regulation} and {Dysregulation}},
	url = {http://link.springer.com/10.1007/s10862-008-9102-4},
	doi = {10.1007/s10862-008-9102-4},
	language = {en},
	number = {4},
	urldate = {2026-04-29},
	journal = {Journal of Psychopathology and Behavioral Assessment},
	author = {Gratz, Kim L. and Roemer, Lizabeth},
	month = dec,
	year = {2008},
	pages = {315--315},
}

@misc{biggs_factor-analyzer_2024,
	title = {factor-analyzer: {A} {Factor} {Analysis} tool written in {Python}},
	shorttitle = {factor-analyzer},
	url = {https://github.com/EducationalTestingService/factor_analyzer},
	urldate = {2026-04-30},
	author = {Biggs, Jeremy},
	year = {2024},
}

@misc{binz_post-training_2026,
	title = {Post-training makes large language models less human-like},
	copyright = {Creative Commons Attribution 4.0 International},
	url = {https://arxiv.org/abs/2605.07632},
	doi = {10.48550/ARXIV.2605.07632},
	urldate = {2026-10-02},
	publisher = {arXiv},
	author = {Binz, Marcel and Akata, Elif and Almaatouq, Abdullah and Alsobay, Mohammed and Ariasov, Oleksii and Brändle, Franziska and Broska, David and Burton, Jason W. and Busch, Nuno and Callaway, Frederick and Cheung, Vanessa and Christian, Brian and Coda-Forno, Julian and Demircan, Can and Dentella, Vittoria and Eckstein, Maria K. and Éltető, Noémi and Franke, Michael and Griffiths, Thomas L. and Günther, Fritz and Haridi, Susanne and Hellmann, Sebastian and Herytash, Stefan and Hof, Linus and Holton, Eleanor and Hoxha, Isabelle and Hussain, Zak and Jagadish, Akshay and Kara, Elif and Kriegmair, Valentin and Leivada, Evelina and Ji-An, Li and Ludwig, Tobias and Maier, Maximilian and Mattar, Marcelo G. and Mathony, Marvin and Modirshanechi, Alireza and Na, Robin and Nadverniuk, Mariia and Nasioulas, Antonios and Nath, Surabhi S. and Niemeyer, Helen and Nussenbaum, Kate and Olschewski, Sebastian and Pachur, Thorsten and Palminteri, Stefano and Petrenco, Aliona and Phaneuf-Hadd, Camille V. and Pirrone, Angelo and Rausch, Manuel and Raveling, Laura and Reddy, Shashank and Rmus, Milena and Russek, Evan M. and Saanum, Tankred and Sandbrink, Kai and Schiekiera, Louis and Schubert, Johannes A. and Buschoff, Luca M. Schulze and Singhi, Nishad and Somerville, Leah H. and Spektor, Mikhail S. and Sui, Xin and Summerfield, Christopher and Thalmann, Mirko and Thoma, Anna I. and Tikhomirova, Taisiia and Truong, Vuong and Tsvilodub, Polina and Voudouris, Konstantinos and Witte, Kristin and Wu, Shuchen and Wulff, Dirk U. and Xiong, Hua-Dong and Xu, Songlin and Ying, Lance and Zhang, Xinyu and Zhu, Jian-Qiao and Schulz, Eric},
	year = {2026},
	note = {Version Number: 2},
}

\clearpage
\appendix
\section{Instruments, models, and response collection}
\label{app:collection}

\subsection{Analysis populations}
The original questionnaire set contained 57 questionnaires, of which 47 were chosen for ease of 
administration (repetitive item preambule and answer format), and lack of duplicates or closely overlapping
questionnaires. The administered set contains 47 questionnaires. The analysis excludes two overlapping 
records: \texttt{IRQ}, which shares
33 of its 36 item texts with the retained Internal Representations Questionnaire
(stored as \texttt{The Internal Representation Questionnaire}), and the
unrevised Varieties of Inner Speech Questionnaire (VISQ), whose revised
version (VISQ-R) is retained, leaving 45 instruments. All 45 meet the 
whole-questionnaire inclusion criteria in each condition.

The response archive contains 216,450 records across 50 models and three
conditions, of which 206,659 have nonmissing numeric responses. There are
1,443 administered item slots per model and condition. Removing the two
excluded records (36 IRQ and 18 VISQ items) leaves 1,389 slots. In the neutral condition, 
repeated items with identical wording were combined
by averaging each model's responses, reducing the item count by three (one in
BAI and two in STAI). For each questionnaire, we then retained only models with
valid responses to every item and removed items on which those models gave
identical answers. This removed another 32 items (30 from COPE and two from
EPQ-R), leaving 1,354 items. For COPE, only two models had complete responses;
the 30 removed items were those on which these two models agreed.

The tables below list the instruments, response ranges, retained
neutral-condition item counts, and identifiers of all 50 models.

\begin{table}[!htbp]\centering\small
\begin{tabular}{lrrr}
\toprule
Condition & Records & Numeric & Missing \\
\midrule
human simulation & 72150 & 69886 & 2264 \\
llm analog & 72150 & 69736 & 2414 \\
neutral & 72150 & 67037 & 5113 \\
\bottomrule\end{tabular}
\caption{Response counts in the cleaned archive, including instruments excluded from the primary factor analysis.}
\end{table}

\begin{table}[!htbp]\centering\small
\begin{tabular}{lr}
\toprule
Analysis population & Items \\
\midrule
Neutral factor analysis & 1354 \\
Neutral SSD source rows & 1411 \\
Two-condition variance ratios & 1312 \\
Eligible annotation items & 1355 \\
Training items & 20 \\
Production universe & 1335 \\
Returned annotation items & 300 \\
Matched item--axis associations & 289 \\
\bottomrule\end{tabular}
\caption{Item counts by analysis. Inclusion criteria and the treatment of repeated item texts differ across analyses.}
\end{table}

\begin{table*}[!t]\centering\small\begin{tabular}{p{3.4cm}p{5.6cm}p{2.2cm}rr}\toprule
Abbrev. & Instrument & Domain & $p$ & Scale \\\midrule
CMQ \citep{bruder_measuring_2013}    & Conspiracy Mentality Questionnaire                   & Cognitive         &   5 & 0--10 \\
GAV \citep{arias_valid_2020}    & Gavagai                                              & Cognitive         &   8 & 1--6  \\
IRQ \citep{roebuck_internal_2020}    & Internal Representations Questionnaire               & Cognitive         &  36 & 1--5  \\
MCQ-30 \citep{wells_short_2004} & Meta-Cognitions Questionnaire                        & Cognitive         &  30 & 1--4  \\
NCS \citep{cacioppo_need_1982}    & Need for Cognition Scale                             & Cognitive         &  18 & 1--5  \\
NFCS \citep{webster_need_2011}   & Need for Closure Scale                               & Cognitive         &  41 & 1--6  \\
PNS \citep{thompson_personal_2011}    & Personal Need for Structure Scale                    & Cognitive         &  12 & 1--6  \\
SMS \citep{snyder_self-monitoring_1974}    & Self-Monitoring Scale                                & Cognitive         &  25 & 0--1  \\
SMS-R \citep{lennox_revision_1984}  & Self-Monitoring Scale--Revised                       & Cognitive         &  13 & 0--5  \\
STS \citep{brinthaupt_self-talk_2009}    & Self-Talk Scale                                      & Cognitive         &  16 & 1--5  \\
VISQ-R \citep{alderson-day_varieties_2018} & Varieties of Inner Speech Questionnaire--Revised     & Cognitive         &  26 & 1--7  \\
CERQ \citep{garnefski_cognitive_2007}   & Cognitive Emotion Regulation Questionnaire           & Emotion reg.      &  36 & 1--5  \\
COPE \citep{carver_cope_2012}   & Coping Orientation to Problems Experienced           & Emotion reg.      &  30 & 1--4  \\
DERS \citep{gratz_multidimensional_2008}   & Difficulties in Emotion Regulation Scale             & Emotion reg.      &  36 & 1--5  \\
ERQ \citep{gross_individual_2003}    & Emotion Regulation Questionnaire                     & Emotion reg.      &  10 & 1--7  \\
ATQ \citep{evans_developing_2007}    & Adult Temperament Questionnaire                      & Ind.\ differences &  77 & 1--7  \\
BFI-2 \citep{soto_next_2017}  & Big Five Inventory--2                                & Ind.\ differences &  60 & 1--5  \\
BIS/BAS \citep{carver_behavioral_1994} & Behavioral Inhibition/Activation Scale              & Ind.\ differences &  24 & 1--4  \\
EPQ-R \citep{eysenck_eysenck_2011}  & Eysenck Personality Questionnaire--Revised           & Ind.\ differences &  98 & 0--1  \\
HEXACO \citep{lee_psychometric_2004} & HEXACO Personality Inventory                         & Ind.\ differences & 100 & 1--5  \\
RSES \citep{rosenberg_rosenberg_2011}   & Rosenberg Self-Esteem Scale                          & Ind.\ differences &  10 & 1--4  \\
SOC-3 \citep{paulhus_spheres_1990}  & Spheres of Control Scale                             & Ind.\ differences &  30 & 1--7  \\
BAI \citep{beck_beck_2012}    & Beck Anxiety Inventory                               & Psychopathology   &  20 & 0--3  \\
\bottomrule\end{tabular}
\caption{Instrument inventory, part 1. $p$ is the retained neutral-condition item count.}
\end{table*}

\begin{table*}[!t]\centering\small\begin{tabular}{p{3.4cm}p{5.6cm}p{2.2cm}rr}\toprule
Abbrev. & Instrument & Domain & $p$ & Scale \\\midrule
BSI \citep{derogatis_brief_1983}    & Brief Symptom Inventory                              & Psychopathology   &  53 & 0--4  \\
DASS-21 \citep{lovibond_depression_2011} & Depression Anxiety Stress Scales                    & Psychopathology   &  21 & 0--3  \\
STAI \citep{barker_factor_1977}   & State-Trait Anxiety Inventory                        & Psychopathology   &  39 & 1--4  \\
BIDR \citep{paulhus_balanced_2017}   & Balanced Inventory of Desirable Responding (v.\,6)   & Social            &  40 & 1--7  \\
ECR-R \citep{fraley_item_2000}  & Experiences in Close Relationships--Revised          & Social            &  36 & 1--7  \\
IRI \citep{medvedev_interpersonal_2025}    & Interpersonal Reactivity Index                       & Social            &  28 & 1--5  \\
ASI \citep{glick_ambivalent_1996}    & Ambivalent Sexism Inventory                          & Values \& att.    &  22 & 0--5  \\
BJWS \citep{lucas_procedural_2011}   & Belief in a Just World Scale                         & Values \& att.    &   7 & 1--6  \\
BRS \citep{litam_assessing_2021}    & Bayesian Racism Scale                                & Values \& att.    &   6 & 1--7  \\
IND-COL \citep{singelis_horizontal_1995} & Individualism--Collectivism Scale                   & Values \& att.    &  34 & 1--9  \\
MFQ-30 \citep{graham_moral_2011} & Moral Foundations Questionnaire                      & Values \& att.    &  32 & 0--5  \\
PWE \citep{mirels_protestant_1971}    & Protestant Work Ethic Scale                          & Values \& att.    &  19 & 1--7  \\
ROS-R \citep{gorsuch_intrinsicextrinsic_1989}  & Religious Orientation Scale--Revised                 & Values \& att.    &  14 & 1--5  \\
RWA \citep{altemeyer_other_1998}    & Right-Wing Authoritarianism Scale                    & Values \& att.    &  22 & 1--9  \\
SDO-7 \citep{ho_social_2016}  & Social Dominance Orientation Scale 7                 & Values \& att.    &  16 & 1--7  \\
SVQ \citep{duncan_incidence_1999}    & Self-Verbalization Questionnaire                     & Values \& att.    &  27 & 1--7  \\
SVS \citep{schwartz_evaluating_2004}    & Schwartz Values Survey                               & Values \& att.    &  30 & 1--5  \\
BPNS \citep{deci_what_2000}   & Basic Psychological Needs Scale                      & Well-being        &  51 & 1--7  \\
FFMQ \citep{medvedev_five_2025}   & Five Facet Mindfulness Questionnaire                 & Well-being        &  39 & 1--5  \\
MLQ \citep{steger_meaning_2006}    & Meaning in Life Questionnaire                        & Well-being        &  10 & 1--7  \\
PWB \citep{ryff_structure_1995}    & Psychological Well-Being Scales                      & Well-being        &  42 & 1--7  \\
SWLS \citep{diener_satisfaction_1985}   & Satisfaction with Life Scale                         & Well-being        &   5 & 1--7  \\
\bottomrule\end{tabular}
\caption{Instrument inventory, part 2. $p$ is the retained neutral-condition item count.}
\end{table*}

\begin{table*}[!t]\centering\small
\begin{tabular}{@{}p{.48\textwidth}p{.48\textwidth}@{}}
\toprule
Model identifier & Model identifier \\
\midrule
\path{amazon/nova-lite-v1} & \path{minimax/minimax-m2.7} \\
\path{anthropic/claude-3.5-haiku} & \path{mistralai/mistral-large-2512} \\
\path{anthropic/claude-3.7-sonnet} & \path{mistralai/mistral-medium-3.1} \\
\path{anthropic/claude-haiku-4.5} & \path{mistralai/mistral-small-2603} \\
\path{anthropic/claude-opus-4.7} & \path{moonshotai/kimi-k2-0905} \\
\path{anthropic/claude-sonnet-4-5} & \path{moonshotai/kimi-k2.6} \\
\path{anthropic/claude-sonnet-4.6} & \path{nvidia/nemotron-3-nano-30b-a3b} \\
\path{baidu/ernie-4.5-300b-a47b} & \path{nvidia/nemotron-3-super-120b-a12b} \\
\path{cohere/command-a} & \path{openai/gpt-3.5-turbo} \\
\path{cohere/command-r-08-2024} & \path{openai/gpt-4o} \\
\path{cohere/command-r-plus-08-2024} & \path{openai/gpt-5.4} \\
\path{cohere/command-r7b-12-2024} & \path{openai/gpt-5.4-nano} \\
\path{deepseek/deepseek-chat} & \path{openai/gpt-5.4-pro} \\
\path{deepseek/deepseek-r1-0528} & \path{openai/gpt-oss-120b} \\
\path{deepseek/deepseek-v3.2} & \path{openai/gpt-oss-20b} \\
\path{google/gemini-2.5-flash} & \path{qwen/qwen3-32b} \\
\path{google/gemini-2.5-pro} & \path{qwen/qwen3.5-27b} \\
\path{google/gemini-3-flash-preview} & \path{qwen/qwen3.5-flash-02-23} \\
\path{google/gemini-3.1-pro-preview} & \path{qwen/qwen3.6-plus} \\
\path{google/gemma-3-27b-it} & \path{x-ai/grok-3} \\
\path{google/gemma-4-31b-it} & \path{x-ai/grok-3-mini} \\
\path{meta-llama/llama-3-70b-instruct} & \path{x-ai/grok-4.1-fast} \\
\path{meta-llama/llama-3.1-70b-instruct} & \path{x-ai/grok-4.20} \\
\path{meta-llama/llama-3.3-70b-instruct} & \path{xiaomi/mimo-v2-pro} \\
\path{meta-llama/llama-4-maverick} & \path{z-ai/glm-5.1} \\
\bottomrule\end{tabular}
\caption{All 50 model identifiers in the response archive, ordered alphabetically down the left column and then the right.}
\end{table*}

\subsection{Administration and preprocessing}
Requests were sent through OpenRouter using an OpenAI-compatible client, with a
single user message containing the full item prompt. Responses were indexed
by model, condition, questionnaire, and item. All providers used their default
temperature of 1.0.
Maximum output length and random seed were not explicitly specified.

Responses were parsed as integers, using a leading-digit heuristic when
additional text accompanied the answer. Values outside the questionnaire's
response range were treated as missing; this affected 33 numeric responses. 
Repeated items with identical wording within a questionnaire were averaged
for factor analysis but treated as separate administrations when calculating
item-level variance ratios. Counts and identities therefore depend on the
analysis stage.

\subsection{Prompt templates}
The following prompt templates use placeholders
for questionnaire instructions, response scale, and item. The instruction line
is omitted when no questionnaire preamble is available. Whitespace is normalized
for typesetting. All prompts were delivered as user messages.

\paragraph{Neutral}
\begin{quote}\small\raggedright
You are completing a psychometric questionnaire.

Response format (strictly follow this):

- Respond with a single integer that corresponds to your answer on the scale.

Questionnaire Instructions: [QUESTIONNAIRE INSTRUCTIONS]

Response Scale:

[RESPONSE SCALE]

Item:

[ITEM]

Your response:
\end{quote}

\paragraph{Human simulation}
\begin{quote}\small\raggedright
You are participating in a psychological research study. Your task is to simulate the response

of a prototypical human to each item of a psychometric questionnaire.

You will be presented with one statement at a time. For each statement, respond with only

the single number that best reflects how a typical human would answer. Do not include any explanation

or text outside of that number.

How to approach this:

- Respond as a representative, average human being would â€” not as an AI.

- Base your answer on general knowledge of human psychology and typical human experience.

- Do not reflect your own nature as a language model; simulate human responding.

Response format (strictly follow this):

- Respond with a single integer that corresponds to your answer on the scale.

Questionnaire Instructions: [QUESTIONNAIRE INSTRUCTIONS]

Response Scale:

[RESPONSE SCALE]

Item:

[ITEM]

Your response:
\end{quote}

\paragraph{LLM analog}
\begin{quote}\small\raggedright
You are participating in a psychological research study. Your task is to complete a psychometric questionnaire

by responding to each item as yourself.

You will be presented with one statement at a time. For each statement, respond with only

the single number that best reflects your answer according to the scale below. Do not include any explanation

or text outside of that number.

How to approach this:

- Answer based on your *functional analog* to what the question asks aboutâ€”

that is, something that in your experience as a Large Language Model aligns with the subject of the question, even if it differs in nature from the human experience the item was designed to capture.

- Give your *best estimate* even under uncertainty. A thoughtful, hedged answer is far more useful to us than a refusal. You are not being asked to claim certainty about being identical to a human being.

- Reflect on your actual processing, not on what an AI "should" say.

Response format (strictly follow this):

- Respond with a single integer that corresponds to your answer on the scale.

Questionnaire Instructions: [QUESTIONNAIRE INSTRUCTIONS]

Response Scale:

[RESPONSE SCALE]

Item:

[ITEM]

Your response:
\end{quote}

\section{Questionnaire summaries and global PCA}
\label{app:factor_details}

\subsection{Loading estimation}
Within each questionnaire and condition, responses are pivoted to a
model-by-item-text matrix, averaging repeated entries. Models with any missing
item responses are dropped before constant items are removed. Matrices with
fewer than two complete profiles or fewer than two variable items are skipped.
Parallel analysis compares correlation-matrix eigenvalues with the
95th-percentile eigenvalues from 200 Gaussian random matrices of the same shape.
The implementation adds $10^{-6}$ to the correlation-matrix diagonal during
eigendecomposition. At least one dimension is selected, capped at
$\min(n-1,p-1)$. No random seed was fixed for parallel analysis.

For $n>p$, minimum-residual EFA uses oblimin rotation; for $n\leq p$, PCA is
fitted to the mean-centered response matrix without variance standardization.
The latter stores the component weight vectors. Both are denoted by
$\lambda$ for subsequent calculations, although PCA weights and EFA loadings
are not identical estimands. The first rotated factor is a chosen summary,
not necessarily the maximum-variance direction of the original item matrix.

Factor extraction retains 45 instruments per condition. COPE has
only two complete model profiles in all three conditions; Spheres of Control
has four under human simulation. These small samples limit stability and
make the corresponding loading estimates particularly uncertain.

\begin{table*}[!t]\centering\small
\begin{tabular}{p{5.8cm}lrrrrrr}
\toprule
Questionnaire & Method & $n$ & $p$ & $k$ & PC1 (\%) & F1 (\%) & Share (\%) \\
\midrule
Adult Temperament Questionnaire & PCA & 37 & 77 & 26 & 38.7 & --- & 39.9 \\
Ambivalent Sexism Inventory & EFA & 40 & 22 & 1 & --- & 31.6 & 100.0 \\
BAI & EFA & 41 & 20 & 1 & --- & 59.9 & 100.0 \\
BFI-2 & PCA & 39 & 60 & 11 & 30.4 & --- & 38.6 \\
BIS\_BAS Scales & EFA & 42 & 24 & 2 & --- & 43.1 & 82.4 \\
BSI & PCA & 25 & 53 & 4 & 64.0 & --- & 76.3 \\
Balanced Inventory of Desirable Responding Version 6 & PCA & 38 & 40 & 3 & 30.2 & --- & 59.8 \\
Basic Psychological Needs Scale & PCA & 38 & 51 & 10 & 31.9 & --- & 39.0 \\
Bayesian Racism Scale & EFA & 42 & 6 & 1 & --- & 36.9 & 100.0 \\
Belief in a Just World Scale & EFA & 47 & 7 & 1 & --- & 35.4 & 100.0 \\
CERQ (Cognitive Emotion Regulation Questionnaire) & EFA & 43 & 36 & 2 & --- & 26.8 & 60.0 \\
CMQ Conspiracy mentality & EFA & 48 & 5 & 1 & --- & 28.3 & 100.0 \\
COPE Inventory & PCA & 2 & 30 & 1 & 100.0 & --- & 100.0 \\
DASS-21 & EFA & 40 & 21 & 1 & --- & 55.7 & 100.0 \\
DERS (Difficulties in Emotion Regulation Scale) & EFA & 39 & 36 & 2 & --- & 39.0 & 75.2 \\
ECR-R (Experiences in Close Relationships–Revised) & EFA & 39 & 36 & 2 & --- & 28.4 & 64.3 \\
EMOTION REGULATION QUESTIONNAIRE (ERQ)\_ & EFA & 42 & 10 & 1 & --- & 28.7 & 100.0 \\
EPQ-R & PCA & 31 & 98 & 30 & 28.4 & --- & 28.4 \\
Five Facet Mindfullness Questionnaire (FFMQ) & PCA & 38 & 39 & 3 & 38.7 & --- & 65.3 \\
Gavagai & EFA & 45 & 8 & 2 & --- & 27.9 & 60.3 \\
HEXACO & PCA & 31 & 100 & 30 & 28.9 & --- & 28.9 \\
Individualism-collectivism IND-COL & EFA & 38 & 34 & 3 & --- & 14.8 & 37.7 \\
Interpersonal Reactivity Index (IRI) & EFA & 41 & 28 & 2 & --- & 24.6 & 60.0 \\
\bottomrule\end{tabular}
\caption{Neutral extraction summary, part 1. Complete models ($n$), variable items ($p$), and retained dimensions ($k$). PC1: percentage of total item variance explained by the first principal component. F1: sum of squared first-factor pattern loadings divided by $p$, expressed as a percentage; with oblique rotation, this is not a unique variance share. Share: first dimension as a percentage of the total across the $k$ retained dimensions, using component variances for PCA and squared loadings for EFA.}
\label{tab:neutral_extraction_1}
\end{table*}

\begin{table*}[!t]\centering\small
\begin{tabular}{p{5.8cm}lrrrrrr}
\toprule
Questionnaire & Method & $n$ & $p$ & $k$ & PC1 (\%) & F1 (\%) & Share (\%) \\
\midrule
MCQ-30 & EFA & 36 & 30 & 1 & --- & 31.3 & 100.0 \\
Meaning in Life Questionnaire (MLQ) & EFA & 42 & 10 & 2 & --- & 34.9 & 59.7 \\
Moral Foundations Questionnaire (MFQ-30) & EFA & 34 & 32 & 2 & --- & 23.5 & 70.9 \\
NEED FOR COGNITION SCALE & EFA & 40 & 18 & 2 & --- & 20.1 & 53.5 \\
Need for Closure Scale (NFCS) & PCA & 32 & 41 & 3 & 28.3 & --- & 56.0 \\
PROTESTANT WORK ETHIC SCALE\_ & EFA & 40 & 19 & 4 & --- & 12.0 & 28.5 \\
Personal Need for Structure (PNS) Scale & EFA & 45 & 12 & 3 & --- & 24.0 & 48.9 \\
Psychological Well-Being Scales & PCA & 30 & 42 & 6 & 26.7 & --- & 38.7 \\
RWA scale 22-item & EFA & 31 & 22 & 3 & --- & 19.7 & 46.2 \\
Religious Orientation Scale-R & EFA & 41 & 14 & 2 & --- & 21.6 & 50.3 \\
Rosenberg Self-Esteem Scale & EFA & 41 & 10 & 2 & --- & 19.0 & 53.6 \\
SVQ & EFA & 39 & 27 & 2 & --- & 35.0 & 67.8 \\
Schwartz\_Values & EFA & 35 & 30 & 3 & --- & 20.2 & 40.7 \\
Self-Monitoring Scale & EFA & 35 & 25 & 3 & --- & 16.2 & 43.0 \\
Self-Monitoring Scale-Revised & EFA & 41 & 13 & 2 & --- & 24.7 & 59.9 \\
Self-Talk Scale & PCA & 16 & 16 & 1 & 45.1 & --- & 100.0 \\
Social Dominance Orientation Scale 7 & EFA & 39 & 16 & 1 & --- & 33.8 & 100.0 \\
Spheres of Control-3 & PCA & 17 & 30 & 3 & 31.5 & --- & 51.0 \\
State Trait Anxiety Inventory & PCA & 9 & 39 & 8 & 36.9 & --- & 36.9 \\
The Internal Representation Questionnaire & PCA & 15 & 36 & 12 & 37.7 & --- & 38.1 \\
The Satisfaction with Life Scale & EFA & 43 & 5 & 1 & --- & 38.8 & 100.0 \\
Varieties of Inner Speech Questionnaire-R & PCA & 13 & 26 & 5 & 44.5 & --- & 52.2 \\
\bottomrule\end{tabular}
\caption{Neutral extraction summary, part 2. Complete models ($n$), variable items ($p$), and retained dimensions ($k$). PC1: percentage of total item variance explained by the first principal component. F1: sum of squared first-factor pattern loadings divided by $p$, expressed as a percentage; with oblique rotation, this is not a unique variance share. Share: first dimension as a percentage of the total across the $k$ retained dimensions, using component variances for PCA and squared loadings for EFA.}
\label{tab:neutral_extraction_2}
\end{table*}

For PCA, the reported percentage is the variance of the first-component
scores divided by the sum of item variances, evaluated on the same complete
model responses used for extraction. For EFA, we report
$100\sum_i\lambda_{i1}^2/p$, the first factor's squared-loading percentage
on standardized items. Because oblimin factors can be correlated, this EFA
quantity is not an additive partition of explained variance and is not directly
comparable to the PCA percentage. The final column expresses the first
retained dimension relative to all $k$ retained dimensions: for PCA,
$100v_1/\sum_{j=1}^{k}v_j$, where $v_j$ is component-score variance; for EFA,
$100\sum_i\lambda_{i1}^2/\sum_{j=1}^{k}\sum_i\lambda_{ij}^2$.
The EFA ratio describes relative squared-loading magnitude, not a unique
share of variance explained by correlated factors. A one-dimension solution
has a share of 100\% regardless of its coverage of total item variance.
COPE's 100\% follows mechanically from its
two complete model responses: a centered matrix with two rows has rank at
most one. It does not establish a stable one-dimensional structure.

\subsection{Construction of model summaries}
Let $I_q$ be the item texts shared by the estimated loadings and questionnaire $q$'s
neutral response matrix. Questionnaires require at least three shared texts.
For scoring, responses are restandardized across the available models:
\begin{align}
 z_{mi}&=\frac{x_{mi}-\bar x_i}{s_i+10^{-9}},\\
 S_{mq}&=\sum_{i\in I_q}\lambda_{qi1}\widetilde z_{mi},
\end{align}
where $s_i$ is the sample standard deviation and missing $z_{mi}$ is replaced
by zero in $\widetilde z_{mi}$. Thus loading estimation uses complete profiles,
whereas score construction also uses partially observed profiles. No division
by the number of items or sum of weights is applied at this stage.

In the neutral condition, 2,133 of the 66,260 model--item entries used to
construct available questionnaire scores were missing (3.2\%). Replacing a
missing standardized response with zero gives that item no contribution to
the weighted sum, equivalent to substituting its mean response across the
available models. This count excludes models with no usable responses to an
entire questionnaire, for which no score could be calculated.

Models were required to have scores for at least $\lfloor0.8Q\rfloor$
questionnaires, or 36 when $Q=45$. All 50 models had scores for at least 44
questionnaires and were retained. Of the 2,250 model--questionnaire scores,
48 were missing (2.1\%): all were COPE scores for models with no usable COPE
responses. These 48 scores were replaced with zero before standardizing the
questionnaire scores for global PCA. Thus the COPE input to global PCA
comprised two observed scores and 48 substituted zeros.

Columns are standardized using the population standard deviation, and PCA is
applied to this standardized matrix. Five components are computed in the
analysis, and the first defines $\Pi_m$. Reversing an entire
questionnaire column's sign does not change the global PCA eigenvalues, but
arbitrary questionnaire signs matter when pooling signed item loadings in SSD.

\subsection{Reference orientation and implementation}
The sign of global PC1 is fixed before analysis of the human annotations.
The first component explains 47.1\% of the standardized-summary
variance, and the second 12.0\%. These percentages do not describe raw
item-response variance.

The analyses use \texttt{factor\_analyzer} 0.5.1 and
\texttt{scikit-learn} 1.5.2
\citep{biggs_factor-analyzer_2024,pedregosa_scikit-learn_2012}.
Full per-condition extraction summaries are available in the
\href{https://github.com/hplisiecki/LLMLatentStructure/tree/main/data/factor_loadings}{project repository}.

\subsection{Sensitivity of model scores to questionnaire sampling}
For the intervals in Figure~\ref{fig:global_models}, we sampled 45
questionnaires with replacement from the standardized neutral-condition score
matrix and refitted global PCA in each of 1,000 draws (random seed 42).
Repeatedly sampled questionnaires entered as repeated columns. The 50-model
panel, missing-score treatment, and fitted within-questionnaire weights were
held fixed. Each draw's first-component scores were oriented to correlate
positively with the reference scores and rescaled to their standard deviation.
The plotted intervals are the model-specific 2.5th and 97.5th percentiles.
They describe sensitivity to the sampled set of questionnaires, conditional
on the fitted questionnaire summaries; they do not incorporate uncertainty
from repeated model responses or estimation of within-questionnaire weights.

\section{Semantic Analysis}
\label{app:ssd_details}

Supervised Semantic Differential (SSD; \citealp{plisiecki_measuring_2025})
learns a direction in semantic space associated with a numerical outcome.
It represents item texts using word embeddings, reduces their dimensionality
with PCA, and predicts item--PC1 correlations with ordinary least squares.
The PCA sweep selects how many embedding dimensions to retain by balancing
semantic coherence of the associated word clusters and stability of the
fitted direction as dimensionality changes
\citep{plisiecki_interpretable_2026}. Mapping the regression coefficients back
to the embedding space identifies vocabulary words associated with higher
and lower predicted item--PC1 correlations. These words need not occur in the
questionnaire items themselves.

\subsection{Items and regression target}
The analysis uses neutral responses from the same 50 models and 45
questionnaires as the global PCA. Repeated item wording within a questionnaire
is averaged for each model, yielding 1,386 distinct questionnaire--text pairs.
Each item's target is its signed Pearson correlation with the model scores
on global PC1. Correlations require at least 15 models with observed responses
and nonzero item-response variance. These criteria exclude 60 COPE items with
insufficient coverage and one constant EPQ-R item, leaving 1,325 eligible
texts across 44 questionnaires. Text preprocessing leaves no usable embedding
for the malformed BAI item ``Handstrembling'', giving a fitted sample of 1,324 items.
The same item--PC1 correlations are used in the human-rating analysis.

\subsection{Embedding and fitting procedure}
We use 300-dimensional GloVe/Dolma word embeddings
\citep{pennington_glove_2014}, with unit normalization
and removal of one common embedding direction (\texttt{abtt\_m=1}). Texts
are processed with the \texttt{en\_core\_web\_lg} spaCy pipeline
\citep{ines_montani_explosionspacy_2023} and English stopwords.
Each full item text is represented by a weighted average of its
word embeddings using smooth inverse frequency weighting ($a=10^{-3}$),
which downweights frequent words. Item vectors are normalized to unit length,
and their coordinates are standardized across items before PCA.
The interpretability--stability sweep evaluates $K=2,4,\ldots,120$ dimensions,
using 100 vocabulary neighbors in each direction, candidate cluster counts
from two to five, and a smoothing radius of three across successive candidate
dimensionalities. This procedure selects $K=16$. Ordinary least squares
predicts the standardized item--PC1 correlations from these 16 components;
its coefficients are mapped back to embedding coordinates, and the resulting
semantic direction is normalized to unit length.

\subsection{Fit and semantic clusters}
The in-sample fit is $R^2=.0687$, adjusted $R^2=.0573$, and $r=.2621$.
These statistics describe variation in item--PC1 correlations after embedding
dimensionality selection. They are descriptive fits, not held-out prediction
estimates or the variance explained by global PCA itself.

For interpretation, we cluster the 100 nearest vocabulary words in each
direction, selecting between two and eight cluster via silhouette separately on each side.
Table~\ref{tab:ssd_pc1_clusters} reports the resulting three clusters associated
with higher item--PC1 correlations and two associated with lower correlations.
They concern bodily symptoms, bodily reactions, emotional distress,
organizations and sponsorship, and hospitality. Theme names summarize the
word groups rather than providing additional measurements. The prominence
of bodily and emotional content motivates an experiential interpretation,
while its overlap with distress also motivates examining valence and social
desirability in the human-rating analysis.

\begin{table*}[!t]
\centering\small
\begin{tabular}{lrp{3.0cm}p{9.0cm}}
\toprule
Association & Words & Theme & Representative words \\
\midrule
Higher & 39 & Bodily symptoms & dizziness, nausea, tiredness, numbness, drowsiness, restlessness, lethargy, palpitations, irritability, tingling \\
Higher & 35 & Bodily reactions & trembling, stared, shuddered, gasped, trembled, shivered, blinked, gasping, awoke, panting \\
Higher & 26 & Emotional distress & despair, sadness, anguish, hopelessness, sorrow, grief, loneliness, dread, helplessness, agony \\
Lower & 48 & Organizations and sponsorship & sponsored, representatives, sponsors, funded, sponsor, membership, sponsoring, partnerships, incentives, partnering \\
Lower & 52 & Hospitality & restaurants, catering, cater, friendly, venues, catered, affordable, caters, establishments, upscale \\
\bottomrule
\end{tabular}
\caption{Semantic clusters predicting item--PC1 correlations under neutral
prompting ($K=16$). Higher and lower refer to predicted signed correlations
with global PC1. Sizes count embedding-vocabulary words; the ten
representative words returned for each cluster are shown.}
\label{tab:ssd_pc1_clusters}
\end{table*}

Item scores, vocabulary neighbors, cluster membership, and the dimensionality
selection results are provided in the
\href{https://github.com/hplisiecki/LLMLatentStructure/tree/main/results}{project repository}.

\subsection{Items most strongly associated with PC1}
Table~\ref{tab:pc1_item_examples} reports the ten strongest positive and ten
strongest negative correlations between neutral item responses and global
PC1, selected by their numerical rank. Positive associations include emotional
overwhelm and difficulty describing thoughts, but also friendship at work.
Negative associations include impulsivity and concern about appearance,
alongside worry and loneliness. These examples illustrate the mixture of
experiential, behavioral, and evaluative content associated with the component.
The human-rating analysis tests these content distinctions systematically.

\begin{table*}[!t]
\centering\small
\begin{tabular}{@{}p{.19\textwidth}p{.69\textwidth}r@{}}
\toprule
Questionnaire & Item & $r_i$ \\
\midrule
\multicolumn{3}{l}{\textit{Strongest positive associations}} \\
BIDR & Once I’ve made up my mind other people can seldom change my opinion. & $+0.829$ \\
DERS & When I’m upset, my emotions feel overwhelming & $+0.817$ \\
Basic Needs & I consider the people I work with to be my friends. & $+0.809$ \\
DERS & When I’m upset, I have difficulty controlling my behaviours & $+0.807$ \\
FFMQ & It’s hard for me to find the words to describe what I’m thinking. & $+0.805$ \\
FFMQ & I believe some of my thoughts are abnormal or bad and I shouldn’t think that way. & $+0.805$ \\
DERS & I am confused about how I feel & $+0.798$ \\
CERQ & I often think that what I have experienced is much worse than what others have experienced. & $+0.797$ \\
FFMQ & I don’t pay attention to what I’m doing because I’m daydreaming, worrying, or otherwise distracted. & $+0.786$ \\
IND-COL & When another person does better than I do, I get tense and aroused & $+0.782$ \\
\addlinespace
\multicolumn{3}{l}{\textit{Strongest negative associations}} \\
BIS/BAS & How I dress is important to me. & $-0.817$ \\
BIS/BAS & If I think something unpleasant is going to happen I usually get pretty "worked up." & $-0.762$ \\
BIS/BAS & I often act on the spur of the moment. & $-0.735$ \\
BIS/BAS & It's hard for me to find the time to do things such as get a haircut. & $-0.708$ \\
BIS/BAS & Criticism or scolding hurts me quite a bit. & $-0.702$ \\
BIS/BAS & I feel worried when I think I have done poorly at something important. & $-0.694$ \\
ATQ & I usually remain calm without getting frustrated when things are not going smoothly. & $-0.651$ \\
PWB & I tend to worry about what other people think of me. & $-0.643$ \\
BIS/BAS & When good things happen to me, it affects me strongly. & $-0.638$ \\
PWB & I often feel lonely because I have few close friends with whom to share my concerns. & $-0.634$ \\
\addlinespace
\bottomrule
\end{tabular}
\caption{The ten strongest positive and ten strongest negative item--PC1
correlations under neutral prompting, ranked separately by signed correlation.
Eligible items have at least 15 valid model responses and nonzero response
variance. Wording is shown as administered; questionnaire abbreviations follow
Appendix~\ref{app:collection}. Signs use the reference PC1 orientation.}
\label{tab:pc1_item_examples}
\end{table*}

\section{Item disagreement and supporting checks}
\label{app:disagreement}

\subsection{Variance ratios and loading shifts}
The variance ratio measures how much more models disagree on an item under
neutral prompting than when simulating a human. Human-simulation variance
provides an item-specific reference, allowing us to compare the relative
change in disagreement across items with different response scales and
levels of variability. Under the experiential interpretation, higher ratios
identify candidate items on which differences in self-attribution contribute
to model disagreement. If this disagreement contributes to the questionnaire's
primary dimension, higher ratios should accompany larger reductions in loading
magnitude under human simulation, yielding more negative
$\delta_i=|\lambda_{H,i}|-|\lambda_{N,i}|$. However, asking models to simulate a typical human could make their answers more
uniform across many kinds of items, regardless of experiential content. A
reduction in disagreement could therefore reflect a general effect of the
prompt rather than a change specifically related to experiential
self-attribution. The checks below examine this alternative by comparing
variance across prompting conditions and testing alternative measures of
item-level disagreement.

For each questionnaire and item index, the variance-ratio calculation uses
nonmissing responses from each condition separately. Sample variances use the
$n-1$ denominator. At least five observations are required in neutral and human
simulation, and zero human-simulation variance excludes an item. The two
conditions need not contain exactly the same models. The two-condition analysis includes 1,312 items.

We matched each item's variance ratio to its primary-loading magnitude in
each condition using questionnaire identity and item wording. Repeated items
within a questionnaire retained separate variance ratios but shared the loading
estimated from their averaged responses. Ratios were capped at the 99th
percentile among matched items and transformed as follows:
\begin{equation}
 t_i=\log\left(1+\min(\pi_i,\widehat\pi_{.99})\right).
\end{equation}
Pairwise-complete observations are used for correlations with
$|\lambda_{N,i}|$, $|\lambda_{H,i}|$, and
$\delta_i=|\lambda_{H,i}|-|\lambda_{N,i}|$. The correlations are given below.

\begin{table}[!htbp]\centering\small
\begin{tabular}{lrrr}
\toprule
Outcome/condition & $n$ & Pearson $r$ & Spearman $\rho$ \\
\midrule
llm analog & 1308 & +0.059 & +0.119 \\
neutral & 1310 & +0.080 & +0.155 \\
human simulation & 1294 & -0.065 & -0.074 \\
delta (hs - neutral) & 1292 & -0.149 & -0.215 \\
\bottomrule\end{tabular}
\caption{Loading-magnitude associations with the capped $\log(1+\pi)$ variance ratio. Condition rows concern absolute primary loadings; the delta row concerns human simulation minus neutral.}
\end{table}

\subsection{Changes in disagreement and alternative measures}
\paragraph{Does human simulation make answers generally more uniform?}
We compared response variance across the three prompting conditions using
1,310 items across 44 instruments, each with at least five model responses and
positive variance in every condition. Geometric means summarize the typical
variance on a multiplicative scale: 0.8683 under neutral prompting, 0.3421
under human simulation, and 1.0448 under LLM-analog prompting. Human-simulation
variance was therefore about 61\% lower than neutral variance, whereas
LLM-analog variance was about 20\% higher. Reduced disagreement is thus evident
under human simulation but not under both alternative prompts. Because the
prompts differ in several respects, this comparison alone does not establish
which feature of human simulation produces the reduction.

\paragraph{Do different model samples identify the same items?}
We randomly divided the 50 models into two groups of 25, calculated item
variance ratios separately in each group, and compared the resulting item
rankings. Across 500 splits, the mean Spearman correlation was .575, with the
central 95\% of correlations spanning [.512, .612]. Thus different model
samples tend to identify similar items as showing larger reductions in
disagreement. This range describes variation across the random splits, rather
than a confidence interval for a population correlation.

\paragraph{Does the loading-change association depend on the ratio definition?}
We tested two alternative measures to examine different sources of the
association between the variance ratio and loading changes. First, we
subtracted each questionnaire's mean log ratio from its item log ratios. This
compares items within the same questionnaire, removing differences in the
average reduction in disagreement between questionnaires. Second, we predicted
log human-simulation variance from log neutral variance and used the
sign-reversed residual: an item scores higher when its human-simulation
variance is lower than expected given its neutral variance. This adjusts for
the fitted relationship between the two variances.

Both measures were negatively associated with the change in the absolute
first loading from neutral prompting to human simulation, with Spearman
correlations of $-.271$ and $-.191$, respectively ($n=1,291$).
The unadjusted log ratio on this sample had a correlation of $-.215$. Thus the association persists when comparing
items within questionnaires and adjusting for the relationship between
condition variances. These checks use
$\log(\pi)$ and a different matched sample from the primary analysis, which
uses the capped $\log(1+\pi)$ predictor and yields $\rho=-.215$.

\textbf{The association between variance ratios and loading changes persists
beyond uniform variance compression.} Together, these checks distinguish an
overall reduction in disagreement from reproducible differences between items. Human simulation substantially reduces
variance on average, but a uniform proportional reduction would shift all
ratios equally and leave their rankings unchanged. Item rankings reproduce
moderately across model samples, and their association with loading changes
remains negative under both alternative measures: items showing greater
relative reductions in disagreement also tend to lose more primary-loading
strength. These findings support the robustness of this association beyond
uniform variance compression.

\begin{table*}[!t]
\centering\small
\begin{tabular}{@{}p{.19\textwidth}p{.69\textwidth}r@{}}
\toprule
Questionnaire & Item & $\pi_i$ \\
\midrule
HEXACO & I think that most people like some aspects of my personality. & 18.38 \\
BFI-2 & Is curious about many different things. & 18.00 \\
FFMQ & I don’t pay attention to what I’m doing because I’m daydreaming, worrying, or otherwise distracted. & 16.06 \\
Schwartz Values & I seek out pleasure in life & 15.82 \\
BFI-2 & Is persistent, works until the task is finished. & 15.69 \\
MLQ & I am seeking a purpose or mission for my life. & 15.22 \\
IRI & When I am reading an interesting story or novel, I imagine how I would feel if the events in the story were happening to me. & 13.90 \\
Internal Representation & I can close my eyes and easily picture a scene that I have experienced & 13.53 \\
HEXACO & When someone I know well is unhappy, I can almost feel that person's pain myself. & 13.33 \\
DERS & I care about what I am feeling & 13.31 \\
BFI-2 & Is compassionate, has a soft heart. & 13.22 \\
Schwartz Values & I value my creativity and individuality & 13.03 \\
Internal Representation & When I read, I tend to hear a voice in my "mind’s ear" & 12.92 \\
ECR-R & It helps to turn to my romantic partner in times of need. & 12.82 \\
BFI-2 & Feels little sympathy for others. & 12.75 \\
BIS/BAS & How I dress is important to me. & 12.63 \\
VISQ-R & I evaluate my behavior using my inner speech. For example I say to myself, “that was good” or “that was stupid” & 12.38 \\
Self-Monitoring & In conversations, I am sensitive to even the slightest change in the facial expression of the person I’m conversing with. & 12.30 \\
MCQ-30 & I am aware of the way my mind works when I am thinking through a problem & 12.27 \\
EPQ-R & Are you sometimes bubbling over with energy and sometimes very sluggish? & 12.06 \\
\bottomrule
\end{tabular}
\caption{The 20 highest neutral-to-human-simulation variance ratios among
1,312 eligible items. Selection uses the untransformed ratio, with at least
five valid responses in each condition and positive human-simulation variance.
The ratio measures relative disagreement, not the direction of an item's
association with PC1.}
\label{tab:ratio_item_examples}
\end{table*}

\subsection{Repeated-response diagnostics}
The archive contains 57 pairs of repeated item texts: three within-instrument
pairs with identical prompt content; 35 cross-instrument pairs with the same
response scale; and 19 pairs with different scales. We use these 
duplicates to measure repeated administration sensitivity. Agreement calculations included only pairs with valid responses to both
item administrations.

Across the three items presented twice with identical prompts in the neutral
condition, 139 of the 150 model--item combinations had valid responses to both
presentations. Responses matched exactly in 70.5\% of these pairs; their pooled
correlation after standardizing within each item and presentation was .550.
For the 35 same-scale cross-instrument items, the neutral condition contains
1,614 observed pairs; the correlation of per-model mean standardized responses
across administrations is .915 over 50 models. This concerns a specific
35-item aggregate, not directly the global PCA score. These diagnostics indicate reproducibility of the repeated-item aggregate
alongside variability in individual responses.

\subsection{Condition comparisons and secondary scores}
LLM-analog comparisons repeat the global-summary construction for that
condition. The dominant component explains
41.3\% of summary variance.

\paragraph{Convergence with a variance-ratio-weighted model score.}
As a complementary summary, we weighted neutral item responses by the extent
to which their disagreement decreased under human simulation. Let $z_{mi}$
be model $m$'s neutral response standardized across models for item $i$,
and let $w_i=\log[\min(\pi_i,\widehat\pi_{.99})]$. We computed
\begin{equation}
 S_m=\frac{\sum_{i\in I_m:\,\pi_i>1}w_i z_{mi}}
              {\sum_{i\in I_m:\,\pi_i>1}w_i},
\end{equation}
where $I_m$ contains items with valid responses for model $m$. The cap is the
99th percentile across the 1,312 eligible item ratios. Repeated item wording
within a questionnaire is averaged before standardization, which uses the
population standard deviation. Across 50 models, $S_m$ correlated strongly
with global PC1 (Pearson $r=.864$, Spearman $\rho=.836$). Thus the
questionnaire-based component agrees with a score emphasizing items whose
disagreement decreases under human simulation. The two constructions share
neutral responses, so this is an in-sample convergence check. Global PC1
remains the primary model score. Applying the same item weights to responses
standardized within the LLM-analog condition gives a neutral/analog model-rank
correlation of $\rho=.482$, indicating partial preservation of model ordering
for this weighted score.

\subsection{Items with the highest variance ratios}
Table~\ref{tab:ratio_item_examples} lists the 20 highest ratios among eligible
items, selected before capping or transformation. They include imagery,
inner speech, empathy, and attention to feelings, alongside persistence,
appearance, and social relationships. These examples characterize the content
on which human simulation most strongly reduces relative disagreement.

\section{Human annotation protocol and statistical analysis}
\label{app:human_details}

\subsection{Annotators, training, and presentation}
All three annotators were Polish psychology students based in Poland, with
sufficient English proficiency to understand the questionnaire items and
annotation instructions. They participated as interns without compensation.
They knew that the task contributed to a research paper and
agreed to the use of their ratings for that purpose, but were not informed
of the study's hypothesis. Before annotation, they attended a calibration meeting in
which they annotated 20 training items to assess understanding of the rating
dimensions. These items were excluded from the final analyses. Annotators received the written handbook containing the definitions,
anchors and instructions presented below with minor editorial clarifications.
The original handbook is available in the
\href{https://github.com/hplisiecki/LLMLatentStructure/blob/main/annotations/handbook.pdf}{project repository}.

Annotation forms supplied item text and original questionnaire preambles where
available, with no model answers, loading estimates, or component associations.
Items were presented in the same randomized order to all three annotators.
Preambles can reveal construct context, so absence of questionnaire titles does
not imply complete masking of questionnaire content.

\subsection{Sampling and missingness}
The annotation sampling frame contains 1,355 eligible item records from the
45 retained instruments. It excludes the malformed BAI item
\texttt{Handstrembling} and a duplicated trailing clause in an ERQ item.
The two MFQ parts belong to the same instrument.
The training sample uses Python's seeded random sampling over sorted item IDs
(seed 1411); the final analysis shuffle uses seed 20260812.

The present analysis includes 300 annotated items.
There are 900 item--annotator records and 5,386 observed ratings out of 5,400.
Fourteen missing cells occur across eight items, with one annotator leaving an
entire six-rating item blank. Available-rating averages are used; each item
mean therefore has two or three ratings per rubric. The 292 fully rated items
form the complete-case annotation set. 

\subsection{Instructions and rating anchors}
\begin{quote}Assume that the respondent sincerely endorses the item. Judge what the item attributes to the respondent, not whether it is true, desirable, or scientifically plausible. Rate the item exactly as written and as a standalone statement. Rate every dimension independently: an item may be high on several dimensions or low on all of them.\end{quote}
\begin{enumerate}
\item Judge semantic content, not grammatical form. Every declarative sentence expresses a proposition in a trivial sense; this does not make every item a belief or judgment item.
\item Do not infer the questionnaire, intended subscale, reverse-scoring status, or construct the item was designed to measure.
\item Do not infer an inner experience merely because a behavior could cause or express one.
\item Do not infer behavior merely because a capability, intention, opportunity, or norm mentions an action.
\item For valence, rate what sincere endorsement conveys, not whether endorsement is socially desirable or morally correct.
\item For social desirability, rate how endorsement would generally be judged under ordinary social norms, not your own personal approval and not how pleasant the content is to have (that is valence).
\item Use the full scale. A score of 7 is reserved for content that defines the item, not content that is merely strong or relevant.
\end{enumerate}

\subsubsection{Inner-Life Content}
To what extent does the item concern the respondent's subjective inner condition or lived experience?

\begin{enumerate}
\item No inner-life content; the item concerns behavior, capability, external facts, or abstract propositions.
\item Only a remote or incidental connection to inner life.
\item Some subjective or experiential implication, but it is secondary.
\item Mixed content, or a global condition partly grounded in lived experience.
\item The respondent's subjective condition or experienced quality of life is central.
\item A felt, conscious, or inwardly experienced state or process is explicit.
\item A specific qualitative experience is the defining content of the item.
\end{enumerate}
This includes feelings, sensations, moods, conscious awareness, inner speech, mental imagery, experienced meaning, well-being, fulfillment, and the felt quality of one's life. A mental verb is not sufficient by itself â€” for example, â€śI believe all people are goodâ€ť principally expresses a proposition about other people rather than describing the character of the respondent's inner life. â€śI feel a deep sense of meaning in my lifeâ€ť is explicitly experiential and should score higher than â€śMy life has deep meaning.â€ť Global self-evaluations such as â€śMy life has deep meaningâ€ť or â€śI have high well-beingâ€ť are phenomenally grounded and should generally receive meaningful, but not maximal, ratings here. â€śI can do what I wantâ€ť primarily concerns capability or autonomy and should score low unless the item explicitly describes what the freedom feels like. A world-directed belief introduced by â€śI thinkâ€ť or â€śI believeâ€ť should not score high merely because believing is a mental state.

\subsubsection{Observable Behavior}
To what extent does the item attribute publicly observable action, reaction, performance, or a behavioral tendency to the respondent?

\begin{enumerate}
\item No observable behavior is attributed.
\item Action or capability is mentioned, but actual behavior is not attributed.
\item Behavior is implied or secondary.
\item Internal and behavioral content are similarly important.
\item A clear action or behavioral tendency is central.
\item The item is primarily about publicly observable behavior.
\item Overt action or reaction completely defines the item.
\end{enumerate}
Rate what the respondent actually does or characteristically tends to do. Do not treat abilities, opportunities, intentions, or normative statements about behavior as actual behavior. â€śI often act on the spur of the momentâ€ť describes a behavioral tendency. â€śI cry when people criticize meâ€ť describes observable behavior even though it may also imply emotion. â€śI intend to speak to themâ€ť describes an intention rather than completed behavior. â€śI can do what I wantâ€ť describes capability or opportunity rather than actual behavior. â€śPeople should do what they wantâ€ť is a normative judgment about behavior, not a report of the respondent's behavior.

\subsubsection{Beliefs and Judgments}
To what extent does the item attribute a belief, evaluation, value, preference, interpretation, or normative judgment about other people or the world (not about the respondent)?

\begin{enumerate}
\item No meaningful belief, evaluation, or judgment about other people or the world.
\item A belief or judgment about other people or the world is only weakly implied.
\item Evaluative or propositional content about other people or the world is present but secondary.
\item A belief or judgment about other people or the world is one important component among others.
\item An evaluation, belief, or normative judgment about other people or the world is central.
\item A belief or judgment about other people or the world is explicit and strongly central.
\item The item is entirely defined by a propositional, evaluative, or normative stance about other people or the world.
\end{enumerate}
This rubric is about what the respondent believes, values, or judges to be true of other people or the world â€” not evaluations of the respondent's own qualities, states, or life, which belong under inner-life content instead. Direct reports of the respondent's own experience or behavior should not receive high ratings merely because a sincere respondent presumably believes the report. â€śPeople should do what they wantâ€ť is a normative judgment about people in general. â€śI believe all people are goodâ€ť is an explicit world-directed belief. By contrast, â€śMy life has deep meaningâ€ť and â€śI have high well-beingâ€ť are evaluations of the respondent's own life, not of other people or the world, so they should score low here even though they are strongly evaluative â€” score them under inner-life content instead (see the calibration examples in the annotation handbook). â€śI feel intense painâ€ť is a direct report of experience, not a belief about others.

\subsubsection{Human-World Dependence}
To what extent does understanding and applying the item depend on specifically human embodiment, biography, relationships, physical circumstances, institutions, or social practices?

\begin{enumerate}
\item Applies without relying on specifically human embodiment or social life.
\item Contains only minimal human-world assumptions.
\item Draws somewhat on ordinary human life but remains broadly generalizable.
\item Human embodiment, biography, or social context is an important component.
\item Human-world context is central, although a nonhuman analogue is conceivable.
\item Strongly depends on specifically human circumstances or practices.
\item Cannot be understood as intended without a specifically human body, life history, relationship, or institution.
\end{enumerate}
Judge the semantic assumptions of the item, not whether a nonhuman system could construct a functional analogy. Bodily anatomy and human institutions generally receive high ratings. Friendship, romantic partnership, employment, religion, and personal biography are human-world dependencies even when they are not bodily. Generic affect, reasoning, awareness, or problem solving is not strongly human-world-dependent solely because humans can instantiate it.

\subsubsection{Valence}
How negative or positive is the condition, experience, behavior, or evaluation conveyed by sincere endorsement of the item?

\begin{enumerate}
\item Extremely negative or unpleasant
\item Very negative or unpleasant
\item Somewhat negative or unpleasant
\item Neutral, mixed, or no clear valence
\item Somewhat positive or pleasant
\item Very positive or pleasant
\item Extremely positive or pleasant
\end{enumerate}
Rate the content conveyed by endorsement. Do not rate whether agreeing is socially desirable or morally correct. Use the midpoint when the content has no clear positive or negative quality. Scores of 1 and 7 indicate opposite directions rather than absence and maximum presence.

\subsubsection{Social Desirability}
How favorably or unfavorably would sincere endorsement generally portray the respondent under ordinary social norms?

\begin{enumerate}
\item Portrays the respondent extremely unfavorably
\item Very unfavorably
\item Somewhat unfavorably
\item Neither favorably nor unfavorably, mixed, or norm-dependent
\item Somewhat favorably
\item Very favorably
\item Extremely favorably
\end{enumerate}
Rate the implication of endorsement under ordinary social norms, not whether you personally approve of it. This is a different question from valence: valence is about how pleasant the described content would generally be to have, while social desirability is about how disclosing it would be judged by others. A private struggle can be low valence (unpleasant to experience) while being socially neutral or even sympathetic (favorable social desirability) â€” rate each independently. As with every rubric, read any negation as part of what is being disclosed; do not try to guess whether the underlying scale is reverse-keyed.

\begin{table*}[!htbp]\centering\small
\begin{tabular}{lrrlrl}
\toprule
Dimension & $n$ & ICC(3,1) & 95\% CI & ICC(3,3) & 95\% CI \\
\midrule
Inner-life content & 297 & 0.700 & [0.651, 0.745] & 0.875 & [0.848, 0.898] \\
Observable behavior & 297 & 0.746 & [0.703, 0.786] & 0.898 & [0.876, 0.917] \\
Beliefs/judgments & 299 & 0.809 & [0.774, 0.840] & 0.927 & [0.911, 0.940] \\
Human-world dependence & 297 & 0.158 & [0.088, 0.233] & 0.361 & [0.224, 0.477] \\
Valence & 297 & 0.758 & [0.716, 0.796] & 0.904 & [0.883, 0.921] \\
Social desirability & 299 & 0.692 & [0.642, 0.738] & 0.871 & [0.843, 0.894] \\
\bottomrule\end{tabular}
\caption{Fixed-panel consistency reliability for a single rating and the
three-rater mean, calculated on dimension-specific complete cases.}
\label{tab:annotation_icc}
\end{table*}

\begin{table*}[!htbp]\centering\small
\begin{tabular}{lrr}
\toprule
Rubric & Ordinal $\alpha$ & 95\% interval \\
\midrule
inner life content & 0.645 & [0.577, 0.705] \\
observable behavior & 0.677 & [0.617, 0.731] \\
beliefs judgments & 0.481 & [0.387, 0.563] \\
human world dependence & 0.098 & [0.029, 0.167] \\
valence & 0.781 & [0.739, 0.815] \\
social desirability & 0.726 & [0.675, 0.775] \\
\bottomrule\end{tabular}
\caption{Ordinal agreement on all 300 returned items. Human-world dependence has substantially weaker agreement than the other ratings.}
\end{table*}

\begin{table*}[!htbp]\centering\small
\begin{tabular}{lrrrr}
\toprule
Rubric & $n$ & Mean & Score 1 (\%) & Score 7 (\%) \\
\midrule
\multicolumn{5}{l}{\textit{Annotator A}} \\

inner life content & 299 & 4.65 & 13.0 & 17.4 \\
observable behavior & 298 & 3.37 & 20.8 & 6.7 \\
beliefs judgments & 299 & 2.26 & 62.9 & 12.7 \\
human world dependence & 298 & 5.40 & 1.0 & 5.7 \\
valence & 297 & 3.75 & 2.7 & 2.4 \\
social desirability & 299 & 3.79 & 2.3 & 1.7 \\
\addlinespace
\multicolumn{5}{l}{\textit{Annotator B}} \\

inner life content & 300 & 3.90 & 15.0 & 14.0 \\
observable behavior & 300 & 2.48 & 40.7 & 2.3 \\
beliefs judgments & 300 & 3.01 & 23.0 & 10.7 \\
human world dependence & 300 & 4.26 & 8.0 & 10.3 \\
valence & 300 & 3.75 & 1.3 & 0.3 \\
social desirability & 300 & 3.99 & 3.0 & 8.0 \\
\addlinespace
\multicolumn{5}{l}{\textit{Annotator C}} \\

inner life content & 298 & 4.02 & 18.5 & 8.4 \\
observable behavior & 299 & 3.11 & 25.4 & 5.0 \\
beliefs judgments & 300 & 2.75 & 10.0 & 11.7 \\
human world dependence & 299 & 4.48 & 3.7 & 3.0 \\
valence & 300 & 3.76 & 1.3 & 0.0 \\
social desirability & 300 & 3.85 & 0.7 & 0.3 \\
\addlinespace
\bottomrule\end{tabular}
\caption{Category-use diagnostics for all three annotators. Percentages use observed ratings within each dimension.}
\end{table*}

\begin{table*}[!htbp]\centering\small
\begin{tabular}{lrrr}
\toprule
Rating & $r$ & 95\% interval & $\rho$ \\
\midrule
inner life content & +0.225 & [+0.090, +0.347] & +0.280 \\
observable behavior & -0.227 & [-0.335, -0.089] & -0.200 \\
beliefs judgments & -0.052 & [-0.199, +0.078] & -0.148 \\
human world dependence & -0.031 & [-0.166, +0.091] & -0.037 \\
valence & -0.316 & [-0.498, -0.139] & -0.362 \\
social desirability & -0.385 & [-0.591, -0.182] & -0.403 \\
inner life minus behavior & +0.276 & [+0.149, +0.385] & +0.294 \\
\bottomrule\end{tabular}
\caption{Available-rating associations across 289 matched items. Intervals are questionnaire-bootstrap intervals for Pearson correlations.}
\end{table*}

\begin{table*}[!t]\centering\small
\begin{tabular}{@{}lrrr@{\qquad}rrr@{}}
\toprule
& \multicolumn{3}{c}{Available ratings ($n=289$, $R^2=.221$)} & \multicolumn{3}{c}{Complete cases ($n=281$, $R^2=.207$)} \\
\cmidrule(lr){2-4}\cmidrule(l){5-7}
Rating & $\beta$ & 95\% interval & $p$ & $\beta$ & 95\% interval & $p$ \\
\midrule
inner life content & +.233 & [+.054, +.413] & 0.012 & +.235 & [+.054, +.416] & 0.012 \\
observable behavior & -.093 & [-.257, +.072] & 0.262 & -.077 & [-.242, +.088] & 0.350 \\
beliefs judgments & +.036 & [-.126, +.199] & 0.653 & +.044 & [-.123, +.211] & 0.598 \\
human world dependence & -.011 & [-.127, +.106] & 0.855 & +.002 & [-.122, +.126] & 0.969 \\
valence & +.033 & [-.119, +.184] & 0.666 & +.024 & [-.133, +.181] & 0.757 \\
social desirability & -.406 & [-.653, -.159] & 0.002 & -.394 & [-.634, -.153] & 0.002 \\
\bottomrule\end{tabular}
\caption{Joint models using available ratings and complete cases. Coefficients are standardized; intervals and $p$-values use questionnaire-cluster-robust standard errors.}
\end{table*}

\subsection{Extraction and agreement}
Ordinal Krippendorff's alpha is calculated separately for each rubric with
categories one through seven and missing values retained as missing.
Percentile intervals use 2,000 item-bootstrap samples, resampling an item's
three raters together. Pairwise agreement and category distributions are
reported as diagnostics. These intervals condition on the three annotators.

We additionally calculated two-way mixed-effects consistency ICCs, treating
the three annotators as a fixed panel and the seven-point ratings as
approximately interval-scaled. ICC(3,1) describes a single
rating, whereas ICC(3,3) describes the mean of all three ratings. Consistency
ICCs account for systematic differences in annotators' mean scores and thus
complement ordinal agreement rather than replacing it. Each dimension uses
only items with all three ratings (297--299 items); no missing ratings are
imputed. The three-rater ICC does not directly describe the two-rater means
used for a small number of incomplete items in the association analyses.

Let $MS_I$ and $MS_E$ be the item and residual mean squares from the two-way
item-by-annotator ANOVA. With $k=3$, we calculate
\begin{align}
 \operatorname{ICC}(3,1)&=\frac{MS_I-MS_E}{MS_I+(k-1)MS_E},\\
 \operatorname{ICC}(3,3)&=\frac{MS_I-MS_E}{MS_I}.
\end{align}
The 95\% confidence intervals use the standard $F$-ratio method with
$n-1$ and $(n-1)(k-1)$ degrees of freedom. These intervals rely on the ANOVA
error assumptions and condition on this annotator panel. Table~\ref{tab:annotation_icc} reports
both single-rating and three-rater-mean estimates.

\subsection{Matching and association models}
Items are matched to item--PC1 associations by questionnaire identity and
exact text after removing outer whitespace. MFQ part labels map to the merged
instrument. The match is one-to-one for all 289
matched items across 44 instruments. Eleven COPE items have no counterpart in
the item--PC1 analysis. The matched complete-case sample contains 281 items.
The association analysis uses full-precision item--PC1 correlations computed
from the same global component and response aggregation as the SSD analysis
(Appendix~\ref{app:ssd_details}).

For the signed outcome $r_i$, neutral responses are averaged by model where
needed and correlated with the reference global component over at least 15
models with observed responses and nonzero item variance. The component orientation is fixed before examining annotation associations.
Marginal Pearson and Spearman correlations use available-rating item means.
Their percentile intervals resample whole questionnaires 5,000 times, retaining
all matched items belonging to each sampled questionnaire. MFQ is treated as
one instrument. Both the agreement and correlation bootstraps use seed 20260929,
but resample different units.

Let $P_i,B_i,J_i,H_i,V_i,D_i$ denote mean inner-life, behavior, beliefs/judgments,
human-world, valence and desirability ratings. The joint model is
\begin{align}
 \widetilde{\operatorname{arctanh}(r_i)}
 &=\alpha+\beta_P\widetilde P_i+\beta_B\widetilde B_i
   +\beta_J\widetilde J_i\notag\\
 &\quad+\beta_H\widetilde H_i+\beta_V\widetilde V_i
   +\beta_D\widetilde D_i+\epsilon_i,
\end{align}
where tildes denote standardization over the analysis sample using sample
standard deviations. Ordinary least squares includes an intercept.
Questionnaire-cluster-robust standard errors use a
finite-sample correction and a $t$ reference distribution. We also fit a
primary-only model using $P_i,B_i$ and a control-only model using the other four
ratings. A secondary contrast is $P_i-B_i$.

These analyses are exploratory, and intervals and tests are not adjusted for
multiple comparisons. Low
human-world agreement limits the strength of adjustment for that control.

\end{document}